\documentclass[sn-mathphys,Numbered]{sn-jnl}

\usepackage{graphicx}%
\usepackage{multirow}%
\usepackage{amsmath,amssymb,amsfonts}%
\usepackage{amsthm}%
\usepackage{mathrsfs}%
\usepackage[title]{appendix}%
\usepackage{xcolor}%
\usepackage{textcomp}%
\usepackage{manyfoot}%
\usepackage{booktabs}%
\usepackage{algorithm}%
\usepackage{algorithmicx}%
\usepackage{algpseudocode}%
\usepackage{listings}%
\usepackage{bm}%
\usepackage[final]{pdfpages}
\usepackage{pgffor}

\DeclareMathOperator*{\argmin}{argmin}

\begin{document}

\title[Article Title]{Single-model uncertainty quantification in neural network potentials does not consistently outperform model ensembles}

\author[1]{\fnm{Aik Rui} \sur{Tan}}

\author[2]{\fnm{Shingo} \sur{Urata}}

\author[3]{\fnm{Samuel} \sur{Goldman}}

\author[1]{\fnm{Johannes C. B.} \sur{Dietschreit}}

\author*[1]{\fnm{Rafael} \sur{G\'omez-Bombarelli}}\email{rafagb@mit.edu}

\affil[1]{\orgdiv{Department of Materials Science and Engineering}, \orgname{Massachusetts Institute of Technology}}

\affil[2]{\orgdiv{Innovative Technology Laboratories}, \orgname{AGC Inc.}}

\affil[3]{\orgdiv{Computational and Systems Biology}, \orgname{Massachusetts Institute of Technology}}

\abstract{
    Neural networks (NNs) often assign high confidence to their predictions, even for points far out-of-distribution, making uncertainty quantification (UQ) a challenge. When they are employed to model interatomic potentials in materials systems, this problem leads to unphysical structures that disrupt simulations, or to biased statistics and dynamics that do not reflect the true physics. Differentiable UQ techniques can find new informative data and drive active learning loops for robust potentials. However, a variety of UQ techniques, including newly developed ones, exist for atomistic simulations and there are no clear guidelines for which are most effective or suitable for a given case.  
    In this work, we examine multiple UQ schemes for improving the robustness of NN interatomic potentials (NNIPs) through active learning. In particular, we compare incumbent ensemble-based methods against strategies that use single, deterministic NNs: mean-variance estimation, deep evidential regression, and Gaussian mixture models. We explore three datasets ranging from in-domain interpolative learning to more extrapolative out-of-domain generalization challenges: rMD17, ammonia inversion, and bulk silica glass. Performance is measured across multiple metrics relating model error to uncertainty.
    Our experiments show that none of the methods consistently outperformed each other across the various metrics. Ensembling remained better at generalization and for NNIP robustness;  MVE only proved effective for in-domain interpolation, while GMM was better out-of-domain; and evidential regression, despite its promise, was not the preferable alternative in any of the cases. More broadly, cost-effective, single deterministic models cannot yet consistently match or outperform ensembling for uncertainty quantification in NNIPs.  
}


\maketitle

\section{Introduction}

Over the last decade, neural networks (NN) have increasingly been deployed to study complex materials systems. In addition to modeling Quantitative Structure Property/Activity Relationship (QSPR/QSAR), NNs have been used extensively to model interatomic potentials in atomistic simulations of materials\cite{Butler2018,Schwalbe-Koda2020,Kocer2022,Behler2021,Yang2019}. When employed to predict the potential energy surfaces (PESes) of materials systems, NN interatomic potentials (NNIPs) recover the accuracy of \textit{ab initio} methods while being orders of magnitude faster, enabling simulations of larger time- and length-scales\cite{Mueller2020,Botu2017}. Furthermore, NNIPs have displayed sufficient flexibility to learn across diverse chemical compositions and structures\cite{Behler2007,Onat2018ImplantedAlloys,Gokcan2022LearningNetworks}. As a result, a wide range of NNIP architectures has been developed\cite{Schutt2018,Schutt2021,Batzner2022E3-equivariantPotentials,Zubatyuk2021,Unke2019,Klicpera2020a} and used for applications in reactive processes\cite{Gastegger2015,ang2021active}, protein design\cite{Wang2020CombiningEnergy,Wang2020b}, solids\cite{Marchand2020MachineAl-Cu,Jakse2022MachinePhenomena}, solid-liquid interfaces\cite{Natarajan2016}, coarse-graining\cite{Morawietz2016b,Ruza2020}, and more.  
Nevertheless, NNIPs remain susceptible to making poor predictions in extrapolative regimes. This often results in unphysical structures or inaccurate thermodynamic predictions during atomistic simulations that can compromise the scientific validity of the study\cite{Fu2022ForcesSimulations,Morrow2022HowPotentials}.

To maximize NNIP robustness and avoid distribution shift, the training data should represent the same ensemble that the simulation will visit. However, since high-quality \textit{ab initio} calculations are computationally expensive, acquiring new data points through exhaustive exploration of the chemical space is intractable. Quantifying model uncertainty and active learning are thus key to training robust NNIPs. A good uncertainty quantification (UQ) method should estimate both the uncertainty arising from measurement noise (i.e., aleatoric uncertainty), and the uncertainty in the prediction due to the model error (i.e., epistemic uncertainty)\cite{Heid2023}. Epistemic uncertainty originates from data scarcity, model architecture limitations, and/or suboptimal model parameter optimization. Since \textit{ab initio}-calculated training data is free from aleatoric error, NNIPs mostly suffer from epistemic uncertainties due to their strong interpolative but weak extrapolative capabilities. Hence, it is important to acquire new training data efficiently using a good UQ metric to facilitate active learning and mitigate epistemic uncertainty\cite{Schwalbe-Koda2021,Shuaibi2020}.
Out of the many UQ schemes, Bayesian neural networks (BNNs), in which the weights and outputs are treated as probability distributions, offer an inherent UQ scheme\cite{Vandermause2020b,Jinnouchi2019}. However, BNNs suffer from scaling problems and are prohibitively expensive for use as interatomic potentials for bulk materials systems. Another architecture commonly used for UQ is the NN ensemble\cite{NIPS2017_9ef2ed4b}. NN ensemble offers a straightforward approach to estimating uncertainties since NN weight initialization, training, and even hyperparameter choices are stochastic. Although NN ensembles are straightforward and can yield good accuracy, the computational cost associated with training and inference for an ensemble of several independent NNs can be very high for large systems. Other uncertainty estimation methods seek to address these bottlenecks by estimating uncertainty with a single, deterministic NN. Such strategies are particularly interesting due to the lower computational cost of training and inference for a single NN model. Furthermore, the predictive uncertainty of single deterministic NNs can be estimated from a single evaluation of the data, eliminating the need for stochastic sampling to generate approximations of the underlying uncertainty functions. UQ strategies using single deterministic NNs can be broadly grouped into antecedent and succedent methods\cite{Gawlikowski2021ANetworks}. In the succedent scheme, the NN has been trained and predictor uncertainty corresponding to the data set is estimated from the feature space\cite{Nandy2018StrategiesChemistry}, latent space\cite{Janet2019}, or gradient matrices of the trained NNs\cite{Oberdiek2018ClassificationInformation,Lee2020GradientsNetworks}. One instance of a succedent method is introduced by Janet et al.\cite{Janet2019} for non-NNIP application and employs Euclidean norm distance in latent features for uncertainty approximation. A more recent development building upon this work involves fitting Gaussian mixture models on the latent space of the NN\cite{Zhu2022FastPotentials}. In antecedent methods, on the other hand, priors are placed on the input data\cite{Nix1994EstimatingDistribution,Amini2020,Soleimany2021} or conditions are placed on the NN's training objective\cite{Amersfoort2020}. The mean-variance estimation method, a regression-prior scheme, is a commonly used antecedent scheme, where a Gaussian prior distribution is placed on the input data and the NN is trained to predict the mean and variance (also taken as the uncertainty)\cite{Nix1994EstimatingDistribution}. Another antecedent method that has been introduced recently is the deep evidential regression, where a high order prior distribution is placed on top of the input data\cite{Amini2020}.

To the best of our knowledge, there exists no uncertainty quantification method that significantly outperforms all others in the setting of NNIPs. In fact, there seems to be a lack of comparison between UQ methods for this task. Different from the prediction of physical properties (QSAR/QSPR), NNIPs both necessitate and rely upon force-matching gradients in addition to fitting of PESes, making UQ for NNIP applications more challenging and unique\cite{Behler2016a,Kulichenko2021TheDynamics}. In this work, we study if UQ methods using single, deterministic NNs can consistently outperform NN ensembles with respect to their capacity to produce a strong ranking of uncertainties while having lower computational costs. In particular, we use mean-variance estimation (MVE), deep evidential regression, and Gaussian mixture models (GMM). We evaluate their performance in ranking uncertainties using multiple metrics across three different data sets, and analyze the improvement of the stability of molecular dynamics simulations via active learning. In general, we find that ensemble-based methods consistently perform well in terms of uncertainty rankings outside of the training data domain and provide the most robust NNIPs. MVE has been shown to perform well mostly in identifying data points within the training domain that corresponds to high errors. Deep evidential regression, offers less accurate epistemic uncertainty prediction, while GMM is more accurate and lightweight than deep evidential regression and MVE. 

\section{Methods}

Given a data set $\mathcal{D}$ containing a collection of tuples ${(x, y)}$, a neural network takes the input, $x$ and learns a hypothesis function to predict the corresponding output $\hat{y}$. In the case of an NNIP, $x$ contains the atomic numbers $\mathbf{Z} \in \mathbb{Z}^n_+$ and nuclear coordinates $\mathbf{r} \in \mathbb{R} ^{n \times 3}$ of a material system with \textit{n} atoms, while $\hat{y}$ denotes its potential energy value, $\hat{E} \in \mathbb{R}$, and/or the energy-conserving forces, $\hat{\mathbf{F}}_{i}$ acting on the atom \textit{i}. The energy-conserving forces, $\hat{\mathbf{F}}_{i}$ are calculated as the negative gradients of the predicted potential energy with respect to the atomic coordinates, $r_i$
\begin{equation} \label{eq:force}
    \hat{\mathbf{F}}_i = -\frac{\partial \hat{E}}{\partial r_i}
\end{equation}
The Polarizable atom interaction Neural Network (PaiNN), an equivariant message-passing NN was used to learn the interatomic potentials\cite{Schutt2021}, given its balance between speed and accuracy. Details of the architecture and training are shown in Section I of the Supporting Information. 

\subsection{Uncertainty Quantification (UQ)}

Four methods of uncertainty quantification (UQ) were compared: ensemble-based uncertainty\cite{NIPS2017_9ef2ed4b}, mean-variance estimation (MVE)\cite{Nix1994EstimatingDistribution}, deep evidential regression\cite{Amini2020}, and Gaussian mixture models (GMM)\cite{Zhu2022FastPotentials}. Since forces are a derivative of energy, they typically exhibit higher variability upon overfitting or outside the training domain and thus describe epistemic uncertainty better than the energy variance. Hence, the uncertainty estimates were evaluated as $\sigma_F$ for all models\cite{Schwalbe-Koda2021}.

\begin{figure}[ht]
    \centering
    \includegraphics[width=\linewidth]{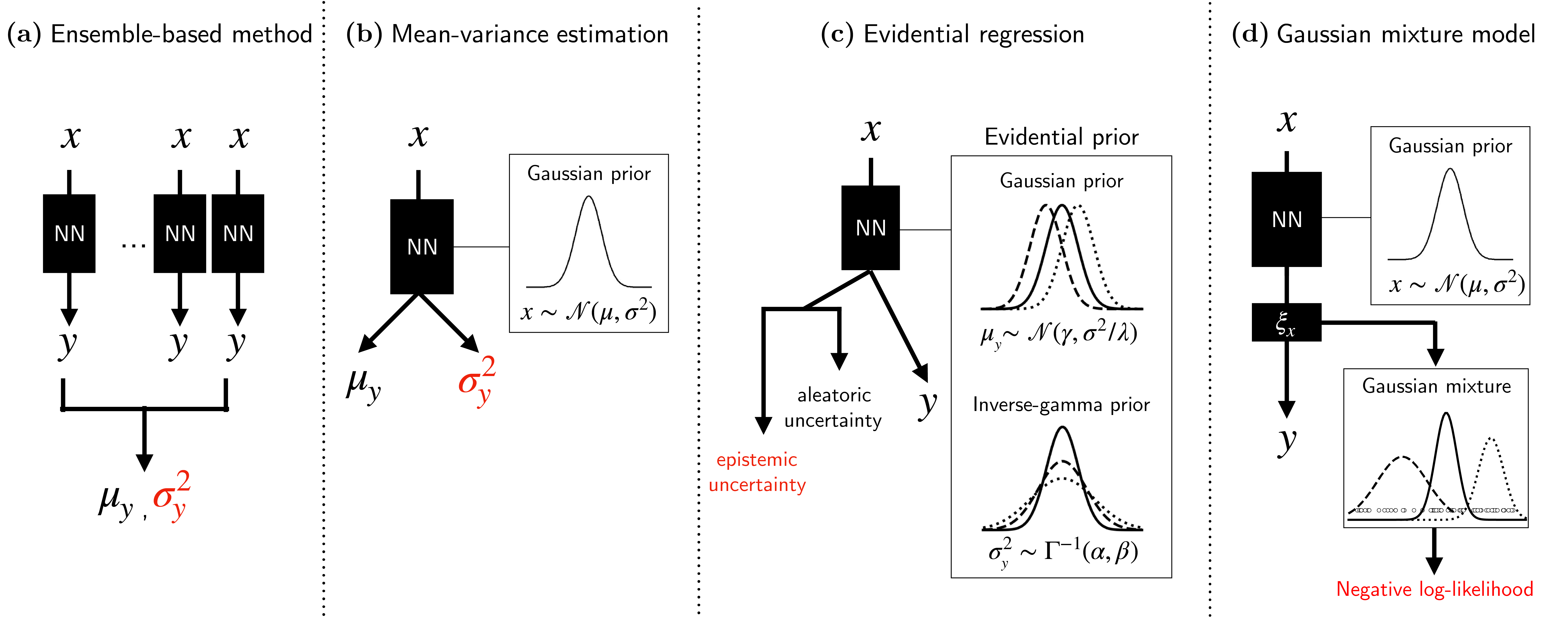}
    \caption{
    Illustration of the uncertainty quantification (UQ) methods used. $x$ denotes the input to the neural networks (NNs), while $y$ is the predicted property. In the case of NNIPs, $x$ generally represents the positions and atomic numbers of the input structure, whereas $y$ is the energy and/or forces of the corresponding input structure. In addition, red texts indicate the variables used as the uncertainty estimates. 
    \textbf{(a)} Multiple NNs trained in an ensemble are made to predict the desired property of the same structure. The mean and variance of the property can then be calculated, where higher variation in the property implies higher uncertainty\cite{NIPS2017_9ef2ed4b,Hirschfeld2020}. 
    \textbf{(b)} In the mean-variance estimation (MVE) method, a Gaussian prior distribution is applied to the input data and the NN is made to predict the mean and variance describing the Gaussian distribution. A higher variance parameter indicates higher uncertainty\cite{Nix1994EstimatingDistribution,Hirschfeld2020}. 
    \textbf{(c)} In the deep evidential regression method, an evidential prior distribution is applied to the input data and the NN predicts the desired property and the parameters to describe both the aleatoric and epistemic uncertainties\cite{Amini2020,Soleimany2021}. 
    \textbf{(d)} In the Gaussian mixture model (GMM) method, the input data is assumed to be drawn from multiple Gaussian distributions. The negative log-likelihood (NLL) function is calculated from a fitted GMM  on the learned feature vectors, $\xi_x$ of the structures. A higher NLL value denotes higher uncertainty\cite{Zhu2022FastPotentials}.}
    \label{fig:method}
\end{figure}

\textbf{Ensemble-based models} are among the most common, cost-effective and trusted approaches to UQ in deep learning. Because there are various sources of randomness in the training of NNs, from initializations to stochastic optimization or hyperparameter choices, a distribution of predicted output values can be obtained from the ensemble-based model that consists of $M$ distinct NNs\cite{NIPS2017_9ef2ed4b,Hirschfeld2020,Gawlikowski2021ANetworks,Schwalbe-Koda2021}. For an input system $x$, each neural network $m$ in the ensemble predicts the total energy, $\hat{E}_m(x)$ and atomic forces $\hat{\mathbf{F}}_m(x)$ (calculated using Eq. \ref{eq:force}). 
The predictions of the ensemble are then taken to be the arithmetic mean of the predictions from each individual NN

\begin{align}
    \mu_E(x) &= \frac{1}{M} \sum^M_m \hat{E}_m(x),
\end{align}

\begin{align}
    \bm{\mu}_F(x) &= \frac{1}{M} \sum^M_m \hat{\mathbf{F}}_m(x)
\end{align}

Lakshminarayanan et al. has proposed that variance of predictions from individual NNs in an ensemble can be used as the uncertainty estimates since the predicted values of NNs may become higher outside of the training data domain\cite{NIPS2017_9ef2ed4b}. In the application of NNIPs, the uncertainty estimates are provided by variances of predicted energy and forces, which can be computed as

\begin{align}
    \sigma_E^2(x) &= \frac{1}{M-1} \sum^M_m (\hat{E}_m(x) - \mu_E(x))^2 
\end{align}

\begin{align}\label{eq:ens_unc}
    \sigma_F^2(x) &= \frac{1}{M-1} \sum^M_m \left[ \frac{1}{3n} \sum_i^n (\hat{\mathbf{F}}_{i, m}(x) - \bm{\mu}_{F_{i}}(x))^2 \right] 
\end{align}

\textbf{Mean-variance estimation (MVE)}. As a function approximator, a standard NN does not have the intrinsic capability to estimate the trustworthiness of its prediction. By treating the training data set as Gaussian random variables with variances, a NN can be constructed to predict the mean, variance pairs that parametrize the Gaussian distributions $\mathcal{N}(\mu, \sigma^2)$\cite{Nix1994EstimatingDistribution}. In other words, the NN predicts a distribution - itself parameterized by a mean and a variance - as an output rather than a fixed point. This method is called mean-variance estimation (MVE), in which the NN predicts the mean and variance for each datum using maximum likelihood estimation. 

The final layer of the NN is modified to predict both the mean, $\mu_{E_i}$ (gradients are subsequently computed to obtain $\mu_{F_i}$) and variance, $\sigma_F^2$ for each datum, and a softplus activation is applied to ensure that the variances are strictly positive. The predicted variances are then taken as the uncertainty metric. Similar to the ensemble method, the uncertainty estimates are described only by the variances of the atomic forces.

\textbf{Deep evidential regression}. While MVE has been widely used for UQ, the predicted variance used in lieu of a proper uncertainty fails to provide estimates of epistemic uncertainties. This is due to the assumption underlying MVE, where the predicted variance of each data point only describes the Gaussian distribution over the data but not over the NN parameters. Thus, MVE does not rigorously inform about distributions outside of the training data regions\cite{Gurevich2020GradientNetworks}. To address this limitation, an emerging evidential regression algorithm has been developed to directly handle representations of epistemic uncertainty by parameterizing a probability distribution function over the likelihood of the learned NN representations\cite{Amini2020}.

Instead of placing priors directly on the data set like MVE, evidential regression places priors on the likelihood function of the data set. It is assumed that the data set is drawn from an independent and identically distributed (i.i.d.) Gaussian distribution ($\mathcal{N}(\cdot)$) with unknown mean and variance ($\mu, \sigma^2$). Here we assume that the data is represented by the atomic forces, $\mathbf{F}_i$. With this assumption, the unknown mean can then be described as a Gaussian distribution while the variance is drawn from an inverse-gamma distribution ($\Gamma(\cdot)$)\cite{Amini2020}. 
\begin{equation}
\begin{split}
    (\mathbf{F}_1, \mathbf{F}_2, ..., \mathbf{F}_N) &\sim \mathcal{N}(\mu_F, \sigma^2_F) \\
    \mu_F \sim \mathcal{N}(\gamma_F, \sigma^2 \nu^{-1}),\ \ &\ \sigma^2_F \sim \Gamma^{-1} (\alpha, \beta)
\end{split}
\end{equation}
where $\gamma$, $\nu$, $\alpha$ and $\beta$ are predicted from the NN, and $\gamma\in\mathbb{R}$, $\nu>0$, $\alpha>1$ and $\beta>0$. Consequently, this higher-order evidential distribution is represented over the mean and variance as 
\begin{equation}
    p(\mu_F, \sigma_F^2 | \gamma_F, \nu, \alpha, \beta) = \frac{\beta^\alpha \sqrt{\nu}}{\Gamma(\alpha)\sqrt{2\pi\sigma_F^2}} \left(\frac{1}{\sigma_F^2}\right)^{\alpha+1} \exp \left( -\frac{2\beta + \nu(\gamma_F - \mu_F)^2}{2\sigma_F^2}\right)
\end{equation}
For detailed derivation, we refer to Ref. \cite{Gurevich2020GradientNetworks,Amini2020}. In a similar fashion to MVE, the final layer of the NN is modified to predict $\gamma_E$ ($\gamma_F$ is then calculated from derivative shown in Eq. \ref{eq:force}), $\nu$, $\alpha$ and $\beta$. A softplus activation is applied to the 3 parameters $\nu$, $\alpha$ and $\beta$ to ensure that they are strictly positive (with a +1 for $\alpha$). Finally, the aleatoric, $\mathbb{E}[\sigma_F^2]$ and epistemic uncertainty, $\text{Var}[\mu_F]$ can be calculated, respectively, as

\begin{align}
    \mathbb{E}[\sigma_F^2] &= \frac{\beta}{\alpha - 1} \\
    \text{Var}[\mu_F] &= \frac{\beta}{\nu (\alpha - 1)}. \label{eq:evi_unc}
\end{align}

\textbf{Gaussian mixture model (GMM)}. A GMM models the distribution of the data set from a weighted mixture of $K$ multivariate Gaussians of dimension $D$\cite{Reynolds2009GaussianBiometrics}. Each $D$-variate Gaussian $k$ in the mixture has a mean vector, $\mathbf{\mu}_k\in\mathbb{R}^{D\times 1}$ and a covariance matrix, $\Sigma_k\in\mathbb{R}^{D\times D}$. The data distribution can then be described as
\begin{equation}
    p(x|{\pi_k, \mu_k, \Sigma_k}) = \sum_{k=1}^K \pi_k \mathcal{N}(\mu_k, \Sigma_k),
\end{equation}
where $\pi_k$ are the mixture weights and satisfy the constraint $\sum_{k=1}^K \pi_k = 1$. To start, the latent features of each training set datum $\xi_{\text{train}} \in \mathbb{R}^{D\times 1}$ are extracted from the immediate layer of a trained NN before the per-atom energy prediction \cite{Zhu2022FastPotentials}. A GMM is constructed on $\xi_{\text{train}}$ using the expectation-maximization (EM) algorithm with full-rank covariance matrices and mean vectors predetermined using k-means clustering. To determine the number of Gaussians needed, we investigated the trade-off between the Bayesian Information Criterion (BIC) and the silhouette score (see Figures~S4 and S5). Using the fitted GMM, we can obtain uncertainty of the test data by taking the latent features, $\xi_{\text{test}}\in \mathbb{R}^{D\times 1}$ and evaluating the negative log-likelihood function $\text{NLL}(\xi_{\text{test}}|\xi_{\text{train}})$ for the test data. For all our models, the dimension of latent features, $D$ were set to 128. A higher $\text{NLL}(\xi_{\text{test}}|\xi_{\text{train}})$ indicates that the data is ``far" from the mean vectors of the Gaussians in the GMM and thus corresponds to high uncertainty.

\begin{equation}
    \text{NLL}(\xi_{\text{test}}|\xi_{\text{train}}) = - \log \left( \sum_{k=1}^K \pi_k \mathcal{N}(\xi_{\text{test}} | \mu_k, \Sigma_k) \right) \label{eq:gmm_unc}
\end{equation}

\subsection{Data Sets}
\textbf{Revised MD17 (rMD17)}. 
The rMD17 data set contains snapshots from long molecular dynamics trajectories of ten small organic molecules\cite{Christensen2020OnForces,Chmiela2017}, in which the energy and forces were calculated at the PBE/def2-SVP level of electronic structure theory. For each molecule, there are 5 splits and each split consists of 1000 training and 1000 testing data points. The data set was obtained from \url{https://doi.org/10.6084/m9.figshare.12672038.v3}. Because the test set is drawn randomly from the same trajectory as the training data, rMD17 is more representative of an in-domain interpolative UQ challenge.

\textbf{Ammonia}.
The ammonia training data set consists of 78 geometries, where the energies and forces are calculated using the BP86-D3/def2-SVP level of theory as implemented in ORCA. The geometries were generated using hessian-displacement in the direction of normal mode vectors on initial molecular conformers generated using RDKit with the MMFF94 force field\cite{Schwalbe-Koda2021}. The data set was obtained from \url{https://doi.org/10.24435/materialscloud:2w-6h}. The test set contains 200 geometries with energies ranging from 0 to 100~kcal/mol (2 geometries in each 1~kcal/mol bin), calculated using the same level theory. The geometries were generated using adversarial sampling during trial runs of this work, and are completely independent of the models and data set analyzed in this work. As the test set comprises higher-energy geometries and the training set encompasses low-energy structures near the ground state, the ammonia dataset serves as a fundamental example of an out-of-domain, extrapolative UQ challenge.

\textbf{Silica glass}.
The silica glass (SiO$_2$) training data set consists of 1590 structures, sampled from multiple long molecular dynamics trajectories subjected to different conditions including temperature equilibration, cooling, uniaxial tension, uniaxial compression, biaxial tension, biaxial compression, and biaxial shear at different rates using the force-matching potential described in Ref \cite{urata2021a}. Each silica glass structure consists of 699 atoms (233 Si and 466 O atoms). In total, 1590 silica structures were sampled from the molecular dynamic trajectories to form the train set and 402 structures were selected using adversarial sampling (obtained during preliminary tests for this study, these samples were acquired separately from the results presented here.\cite{Tan2023data}) to form the test set. DFT calculations were performed on the structures using the Vienna Ab-initio Simulation Package (VASP)\cite{Kresse1996,Kresse1996a,Kresse1999,Perdew1996}. Details of the MD simulations\cite{plimpton1995,Evans1983}, DFT calculations, and adversarial sampling of the silica structures are discussed in Section II.A of the Supporting Information. Because of its high configurational complexity - including high energy fracture geometries - and low chemical complexity, our silica dataset represents a step-up in generalization of UQ in out-of-domain extrapolative regimes.

\subsection{Evaluation Metrics}
\textbf{Spearman's rank correlation coefficient}. Since UQ should measure how much the model does not know about the data presented to it, we expect that a high true error should correlate with a high predicted uncertainty. In other words, we expect a monotonic relationship between the predicted uncertainty, $U$, and the true error, $\epsilon$, for a good uncertainty estimator\cite{Hirschfeld2020,Varivoda2022MaterialsStudy}. Since we are particularly interested in UQ in the context of active learning to improve NNIPs where they are most erroneous, we use the Spearman's rank correlation coefficient to assess the reliability of the uncertainty estimator such that the correlation between the rank variable of the predicted uncertainty, $R_U$, and that of the true error, $R_\epsilon$, can be quantified. The correlation coefficient can then be defined as 

\begin{equation}
    \rho(U, \epsilon) = \frac{\text{cov}(R_U, R_\epsilon)}{\sigma_{R_U} \sigma_{R_\epsilon}}
\end{equation}

\noindent where $\text{cov}(R_U, R_\epsilon)$ is the covariance of the rank variables while $\sigma_{R_U}$ and $\sigma_{R_\epsilon}$ are the standard deviations of the rank variables. In the case of a perfect monotone function, a Spearman correlation of +1 or -1 is expected, whereas a correlation of 0 indicates no association between the ranks of the two variables. However, since this metric only compares ranks of variables, it is still possible for an estimator constantly predicting low uncertainties in the event of high errors to achieve high Spearman coefficient. Hence, complementing this with another metric allows us to understand the performance better.

\textbf{Area under the receiver operating characteristic curve (ROC-AUC)}.
Since uncertainty estimates are expected to be high at high true error points, a criterion to split uncertainties and errors into high and low values is established\cite{Kahle2022QualityEnsembles,Zhu2022FastPotentials}. More precisely, an error threshold, $\epsilon_{c}$ is specified such that data with true error higher than the error threshold, $\epsilon > \epsilon_{c}$ are classified as ``high error" points while data with lower true error, $\epsilon \leq \epsilon_c$ are classified as ``low error" points. Similarly, a specified uncertainty threshold is applied to uncertainty $U_c$. For a structure to be considered as a true positive (TP), both the true error and estimated uncertainty have to be above their corresponding thresholds ($(\epsilon > \epsilon_c) \wedge (U > U_c)$). A false positive (FP) point is a structure with a true error below its threshold while the estimated uncertainty is above the threshold ($(\epsilon \leq \epsilon_c) \wedge (U > U_c)$). Conversely, a true negative (TN) point occurs when $(\epsilon \leq \epsilon_c) \wedge (U \leq U_c)$, and a false negative (FN) when $(\epsilon > \epsilon_c) \wedge (U \leq U_c)$. 

However, setting the threshold values requires manual optimization for each uncertainty method, especially since the magnitude and distribution of uncertainties for each method compared in this work vary extensively. Instead of evaluating the true positive rate (TPR) and predictive positive value (PPV) like in previous work\cite{Kahle2022QualityEnsembles,Zhu2022FastPotentials}, we compute the area under curve (AUC) of the receiver operating characteristic (ROC) curve such that the quality of the uncertainty methods can be assessed independent of the discrimination threshold\cite{Amersfoort2020}. Note that the threshold for ``target" score, which in this case is the threshold for true error, $\epsilon_c$, still has to be specified in order for the ROC curve to be plotted. For all models, we set $\epsilon_c$ to be at the 20th percentile of the error distribution because we find that varying the error percentile threshold does not affect the ROC-AUC score by a significant amount (see Table S2). We also found that using system-dependent absolute cutoff values give the same trend as using percentile cutoffs. The more accurate the method is, the higher the ROC-AUC score. An ROC-AUC score of 0.5 indicates that the method has no discrimination power between high or low uncertainty samples. A score lower than 0.5 implies that the method is better at inverting the uncertainty prediction, predicting low uncertainty for high error points.

\textbf{Miscalibration area.}
Another way to evaluate the quality of the uncertainty estimates is to quantify the calibration error of the models, which is a numerical measure of the accuracy of calibration curve. A calibration curve shows the relationship between the predicted and observed frequencies of data points within different ranges. Tran et al. \cite{Tran2020} proposed to compare how well the expected fraction of errors for each data point falling within $z$ standard deviations of the mean follow the observed Gaussian random variables constructed with the uncertainty estimates ($U(x)$) as the variances. In other words, if the quality of uncertainty estimate is good, we would expect that there will be around 68\% of the errors falling within one standard deviation of the mean of the Gaussian distributions constructed using the uncertainty estimates\cite{Tran2020, Hirschfeld2020,Hu2022,Varivoda2022MaterialsStudy}. From this calibration curve, we can calculate the area between this curve and the line of perfect calibration to provide a single numerical summary of the calibration. This quantity is called the miscalibration area. A perfect UQ method will show an exact agreement between the observed and expected fractions and thus yield a miscalibration area of 0. 

\textbf{Calibrated negative log-likelihood function (cNLL)}.
For observed (true) errors, a negative log likelihood can be calculated as an evaluation metric assuming that the errors are normally distributed with a mean of 0 and variances given by the predicted uncertainties. However, given that some methods do not directly produce uncertainty estimates equivalent to variances, the uncertainty estimates can be calibrated such that they resemble variances more closely\cite{Hirschfeld2020,Janet2019}. Specifically, the ``variances" can be approximated as $\hat{\sigma}^2(x) = a U(x) + b$, where the scalars can be $a$ and $b$ can be estimated as values that minimize the negative log-likelihood function of errors in the validation set, $\mathcal{D}_{val}$,
\begin{equation}
    a_*, b_* = \argmin_{a, b}\ \frac{1}{2} \sum_{x, \mathbf{F} \in \mathcal{D}_{val}} \ln{(2 \pi)} + \ln{(aU(x)+b)} + \frac{(\hat{\mathbf{F}} - \mathbf{F})^2}{(aU(x) + b)} \ .
\end{equation}
With the calibrated scalars $a$ and $b$, the cNLL metric can then be computed on the test set, $\mathcal{D}_{test}$, as
\begin{equation}
    \text{cNLL} = \frac{1}{2} \sum_{x, \mathbf{F} \in \mathcal{D}_{test}} \ln{(2 \pi)} + \ln{(a_*U(x)+b_*)} + \frac{(\hat{\mathbf{F}} - \mathbf{F})^2}{(a_*U(x) + b_*)} \ .
\end{equation}
\subsection{Adversarial Sampling}
In addition to evaluating the reliability of an uncertainty estimator using metrics like Spearman's rank correlation coefficient and ROC-AUC, we would like to assess whether new structures sampled using the estimated uncertainties from each method help to increase robustness of the NNIPs in different ways. To this end, we employed the adversarial sampling method proposed by Schwalbe-Koda et al.\cite{Schwalbe-Koda2021} to sample new conformations that maximize the estimated uncertainty such that these adversarial examples can be incorporated into the retraining of the NNs in an active learning (AL) loop. This would test reliability, smoothness, and utility  of the UQ methods because adversarial sampling requires the UQ method to systematically identify underrepresented regions where the models are most uncertain about. Effectively, improvement in robustness of NNIPs after active learning is a better metric for uncertainty estimation methods since ideally a good uncertainty estimator is able to acquire new training examples that improve the model's generalizability more than random sampling. 

In the adversarial sampling strategy, a perturbation, $\delta$ is applied to randomly chosen structures, $x$ from the initial training data. The perturbation is then iteratively updated using gradient-ascent to maximize the uncertainty of the perturbed structures, $U(x_\delta)$, as defined below
\begin{align}
    \max_{\delta} p(x_\delta) U(x_\delta). \label{eq:thermo_likelihood}
\end{align}
More precisely, $U(x_\delta)$ is taken to be $\sigma^2_F(x_\delta)$ for the ensemble-based method (Eqs. \ref{eq:ens_unc}) and MVE, $\text{Var}[\mu_F(x_\delta)]$ for evidential regression (Eq. \ref{eq:evi_unc}), and $\text{NLL}(x_\delta|x_\text{train})$ for GMM (Eq. \ref{eq:gmm_unc}). $p(x_\delta)$ is the thermodynamic probability of the perturbed structures based on the distribution of data already available to the model 
\begin{align}
    p(x_\delta) = \frac{\exp \left( -\frac{\hat{E}(x_\delta)}{k_\mathrm{B}T} \right) }{\sum_{x\in\text{train}} \exp \left( -\frac{E(x)}{k_\mathrm{B}T} \right) }
\end{align}
where $k_\mathrm{B}$ is the Boltzmann constant, $T$ the absolute temperature, and $\hat{E}(x_\delta)$ in the numerator is the predicted energy of the perturbed structure, while $E(x)$ in the denominator denotes ground truth energy of structures in the initial training data set. Refer to Section~III in the Supporting Information for parameters used for adversarial sampling.

\subsection{Molecular Dynamics (MD) Simulations.}
NN-based MD simulations were performed in the NVT ensemble with the Nos\'e-Hoover thermostat. In the case of ammonia, 100 5~ps-long MD simulations were run for four different temperatures (300~K, 500~K, 750~K, and 1000~K) and for each UQ method at a timestep of 0.5~fs. The initial configurations for all simulations were picked randomly from the initial training data set containing 78 geometries. The trajectories were considered unstable if the distance between the atoms were closer than 0.75~\AA\ or larger than 2.25~\AA\ or if the predicted energy was lower than the ground state energy, which is at a reference point of 0~kcal/mol in this work. For simulations of silica, 10 1~ns-long trajectories at a temperature of 2500~K were performed for each UQ methods at all generations with a time step of 0.25~fs. The initial structures for all simulations were picked randomly from the structures sampled at 2500~K equilibration simulation in the initial training data set. The trajectories were considered unstable if the kinetic energy becomes 0, a null value or greater than 10,000~kcal/mol (due to destabilization of the predicted potential energy in a false minimum\cite{Morrow2022HowPotentials}), or if the distances between any atoms are less than 1.0~\AA. No upper bound was imposed on the atomic distances because the simulations tend to fail before any stretching of bond occurs. All simulations were performed using the Atomic Simulation Environment (ASE) library in the Python language\cite{HjorthLarsen2017}.

\section{Results and Discussion}

\subsection{Revised MD17 Data Set (rMD17)}

\begin{figure}[ht]
    \centering
    \includegraphics[width=\linewidth]{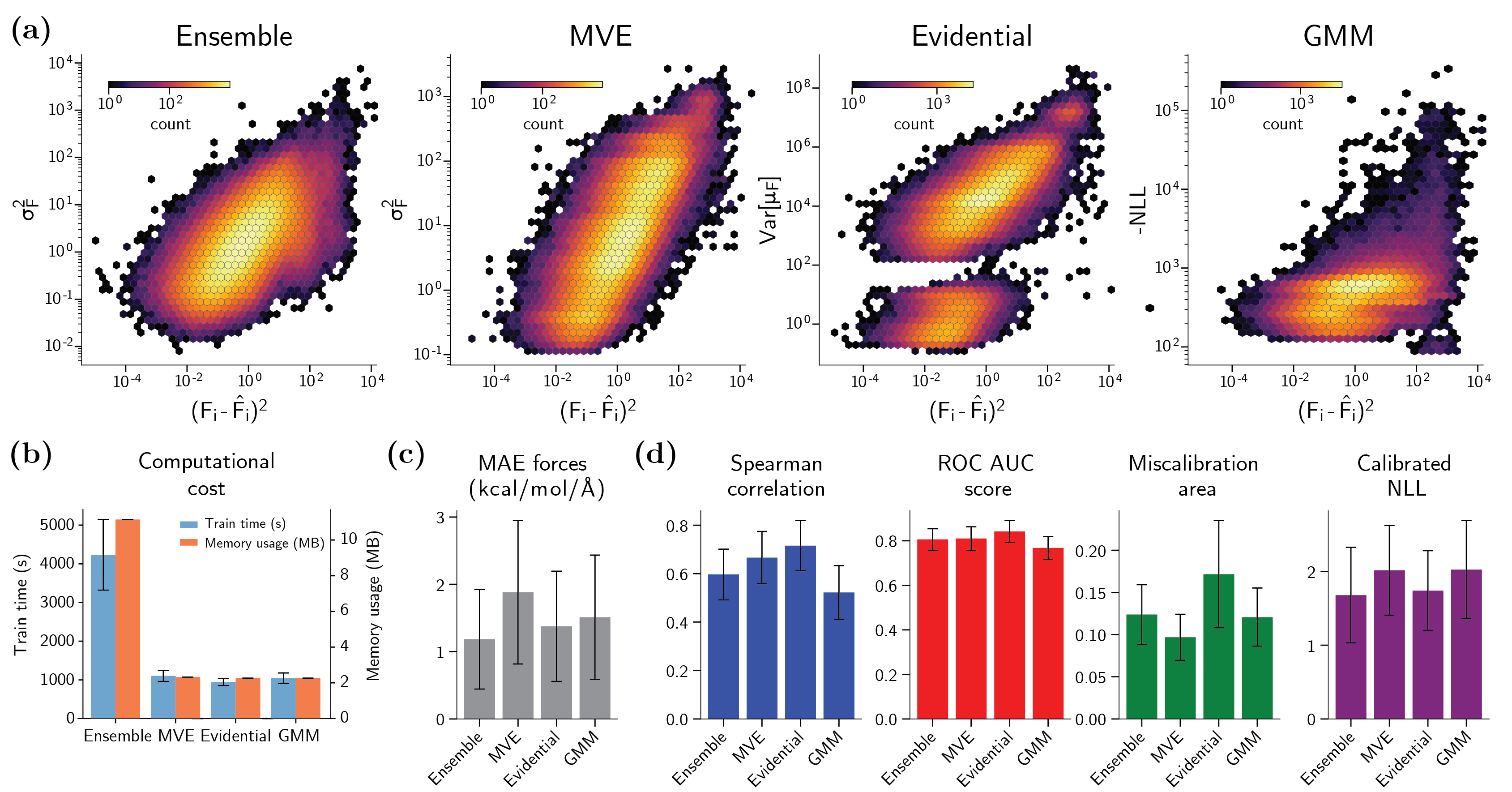}
    \caption{
    \textbf{(a)} Hexbin plots showing (predicted) uncertainties versus squared errors of atomic forces in all molecules of all 5-fold test sets from the rMD17 data set for each considered UQ method. Note that since the uncertainties (-NLL) for the GMM method contain negative values, all uncertainties are scaled by the minimum value to yield positive quantities for plotting of the log scale figure. 
    \textbf{(b)} Total wall time used for training the NNs and total GPU memory usage incurred by the NNs.
    \textbf{(c)} Bar chart showing the prediction errors on the test set of the considered UQ models trained using their corresponding loss functions. Each bar presents the overall average of the metrics for all molecules in all 5-fold test sets of the rMD17 data set. Prediction error is described by the mean absolute error (MAE) of the atomic forces in the unit kcal/mol/\AA.
    \textbf{(d)} Bar charts showing the evaluation metrics for the considered UQ methods, including Spearman's rank correlation coefficient, ROC AUC score, miscalibration area and calibrated negative log-likelihood (NLL) score. 
    For \textbf{(c)} and \textbf{(d)}, refer to Figure~S2 for breakdown of the metrics for each molecule. }
    \label{fig:rmd17}
\end{figure}

As a starting point, we performed UQ on the rMD17 data set, which is readily available online. We evaluated whether the predicted uncertainties are ranked correctly with respect to the true error for all molecules in the data set (Figure~\ref{fig:rmd17}a). An immediate observation is that the uncertainty-error distributions for the ensemble-based and mean-variance estimation (MVE) methods follow a positively-correlated relationship desirable for good UQ schemes. Similarly, the predicted uncertainty of Gaussian mixture model (GMM) exhibits a relatively linear but slightly skewed distribution with respect to the true errors. Evidential regression, on the other hand, shows a two-tiered distribution that does not show good correlation with energy. In addition to that, it is interesting to note that the predicted uncertainties by evidential regression span eight orders of magnitudes, which is due to optimization of the $\nu$, $\alpha$, and $\beta$ parameters during training such that they approach 0. In terms of computational resources used for each method (Figure~\ref{fig:rmd17}b), the training time and GPU memory usage for the ensemble-based method containing 5 independent neural networks (NNs) are roughly 5 times greater than the other single deterministic NN methods. 

Since UQ is often utilized in the context of NNIPs to improve robustness for atomistic simulations, it is critical to first assess the prediction accuracy of each model. From Figure~\ref{fig:rmd17}c, we can see that the ensemble-based method has a lower overall error at inference than the other methods which all employ only a single NN. Given that the choice of model architecture, input representation, and training data for all UQ methods in the same splits are the same, we can assume that all models possess the same model uncertainty (also referred to as model bias), which is the component of epistemic uncertainty associated with model form insufficiencies and lack of data coverage. This suggests that the accuracies of models differ due to suboptimal parameter optimization mainly arising from the difference in loss functions used during training. It has been shown that ensemble learning removes parametric uncertainty (also referred to as model variance) associated with ambiguities in parameter optimization\cite{Gawlikowski2021ANetworks,Heid2023,Gabriel2021UncertaintyReview}, and thus improves generalization which reduces the overall error\cite{Schwalbe-Koda2021,Tumer1996ErrorClassifiers,Mendes-Moreira2012EnsembleSurvey}. MVE, on the other hand, shows the highest average test set error, possibly due to a harder-to-optimize NLL loss function, which has been reported in Ref \cite{Seitzer2022}. Since the robustness of NNIPs largely depends on the accuracy of the models, the overall error is an important criterion to consider while choosing UQ methods to be incorporated into the active learning workflows. In terms of quality of uncertainty estimates (Figure~\ref{fig:rmd17}d), no single method consistently outperforms the other. For the ensemble-based method, all the evaluated metrics are, by comparison, in good ranges. Interestingly, MVE demonstrates good performance in uncertainty ranking despite having low prediction accuracy. Evidential regression, in contrast, outperforms the others in all evaluated metrics except for the miscalibration area, possibly due to the bimodal distribution in Figure~\ref{fig:rmd17}a. Lastly, GMM shows the worst performance in all metrics although within the error bar of the other approaches. We also found that the performance of GMM is independent of the dimension of the latent feature vector and the number of Gaussians (see Figure~S4).

\subsection{Ammonia}

\begin{figure}[ht]
    \centering
    \includegraphics[width=\linewidth]{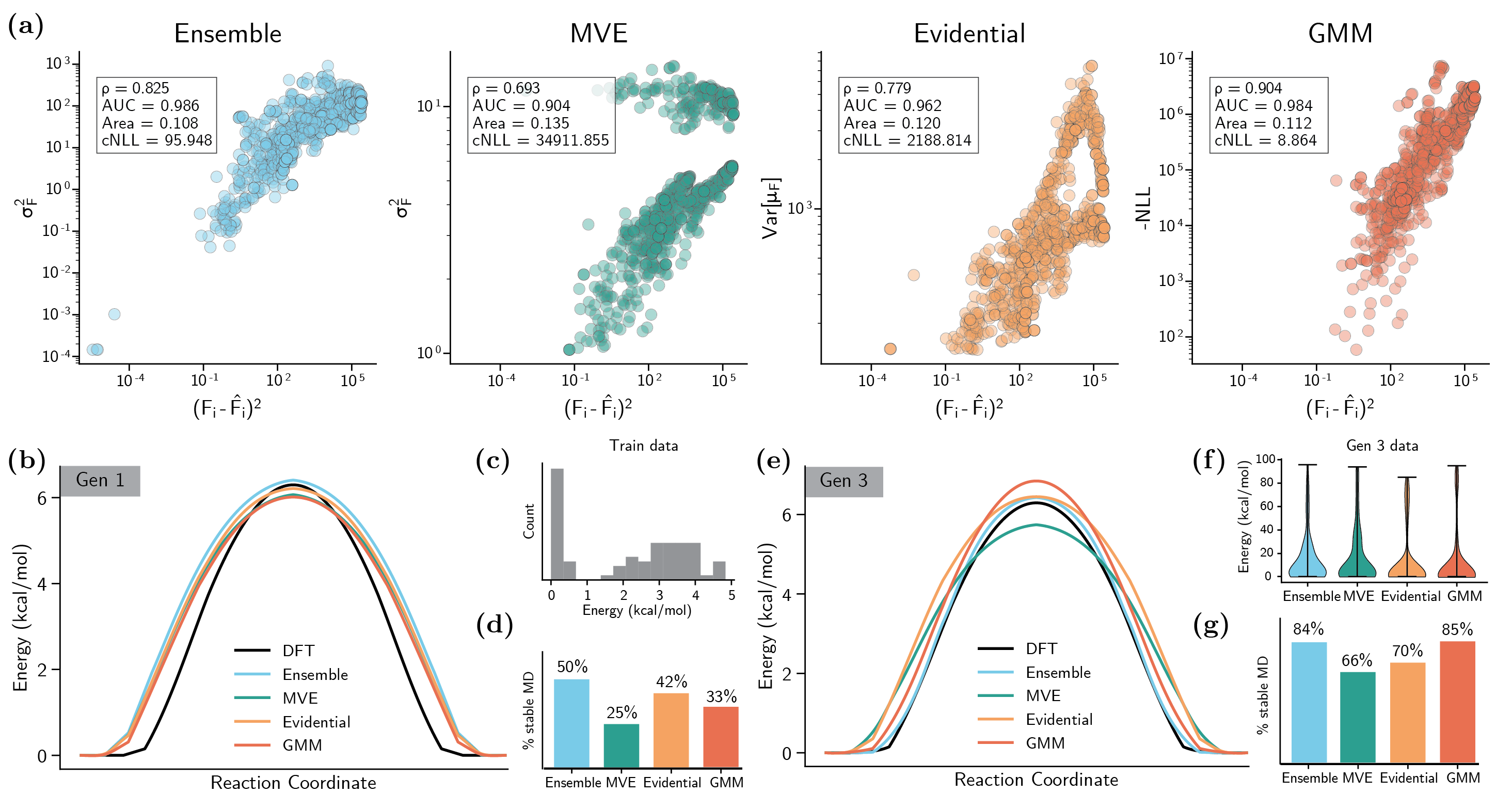}
    \caption{
    \textbf{(a)} Plots showing predicted uncertainties versus squared errors of atomic forces in the test data set for each considered UQ method trained only on the initial training data set (subfigure \textbf{(c)} below). The box at the top right corner of each subfigure shows the evaluation metrics of the methods, namely Spearman's rank correlation coefficient ($\rho$), ROC AUC score (AUC), miscalibration area (Area), and calibrated negative log-likelihoods (cNLL). 
    \textbf{(b)} Energy barrier of nitrogen inversion calculated with NEB using DFT and the considered UQ methods in generation 1. All NNs in the methods were trained on the same initial training data. 
    \textbf{(c)} Histogram showing distribution of energy of geometries in the initial training data. 
    \textbf{(d)} Fraction of stable MD trajectories generated using the NNs of the UQ methods in generation 1 as force field.
    \textbf{(e)} Energy barrier of nitrogen inversion calculated with NEB using DFT and the UQ methods in generation 3. The NNs are trained on new adversarial examples generated with their respective UQ method on top of the initial training data. 
    \textbf{(f)} Distribution of energy in training set after two rounds of adversarial sampling is performed. 
    \textbf{(g)} Fraction of stable MD trajectories generated using the NNs of the UQ methods in generation 3 as force field.
    }
    \label{fig:ammonia}
\end{figure}

In this example, we tested the UQ methods by studying the process of nitrogen inversion in ammonia. In the first generation, all NNs are trained on the same initial training data that contains only 78 geometries with energies ranging between 0 and 5~kcal/mol (Figure~\ref{fig:ammonia}c). When tested on 200 conformations with energies between 0 and 100~kcal/mol (2 conformations per 1~kcal/mol bin), it can be clearly observed that the ensemble-based and GMM model produce good monotonous relationship for the predicted uncertainty with respect to the true error (Figure~\ref{fig:ammonia}a). This can also be confirmed by examining the evaluation metrics for ensemble and GMM in which all metrics outperform the other two methods. Intriguingly, the performance of GMM on this data set is quite different from the results for the rMD17 data set, where MVE and evidential regression discernibly outperform GMM (Figure~\ref{fig:rmd17}d). The most probable reason for this is the difference in distribution of train and test sets. More specifically, the energies of conformations in the train and test sets overlap in the rMD17 data set (Figure~S7), whereas for ammonia, the train set contains only low energy conformations (0 - 5~kcal/mol) while the test set includes conformations in both low and high energy ranges (0 - 100~kcal/mol). This suggests that the ensemble-based and GMM methods are better at predicting epistemic uncertainties outside of the training data domain. Evidential regression is also good at ranking uncertainties with respect to the true errors, but the distribution again shows a peculiar triangular trend. By contrast, MVE underperforms compared to the other models in all evaluated metrics and produces a two-tiered distributions. Comparing the performance of MVE in this data set against the rMD17 data set, we can infer that MVE is likely worse at predicting uncertainties for data outside of the training domain.

When it comes to predicting the nitrogen inversion energy barrier, all the methods seem to either under- or over-estimate the overall ground truth energy curve (Figure~\ref{fig:ammonia}b). This observation is reasonable since all models have only seen low energy conformations in the initial (first generation) training data (Figure~\ref{fig:ammonia}c). When the different models are used as interatomic potentials for MD simulations, NN-ensembles produced the highest fraction of stable MD trajectories at 50\%, in which there are no atomic dissociations or collisions (Figure~\ref{fig:ammonia}d). Evidential regression produces 42\% of stable trajectories, followed by GMM at 33\% and lastly, MVE at 25\%. 

After performing adversarial sampling for two generations, 40 data points were added to training of the NNIPs for each UQ method. It is important to note that the new data points were generated independently by each UQ method and differ from one method to the other. The energy distributions of the new training data for each method, including the initial training data that contains 78 geometries, are shown in Figure~\ref{fig:ammonia}f. Looking closely at the energy distributions, we can see that while the ensemble-based method are able to drive some conformations to higher energies, the majority of the new data points are still concentrated in the lower energy region around 0-30 kcal/mol. This is also similar in the MVE method, but the number of new conformations with energies in the 20-60~kcal/mol range is higher than in the ensemble-based method. In contrast, evidential regression samples more new geometries with energies in the range of 40-85~kcal/mol and below 20~kcal/mol. This may be due to the diverging trend shown in Figure~\ref{fig:ammonia}a, where the maximization of uncertainty during the adversarial sampling pushes geometries to low and high energy ranges in a tiered manner. Similar observations can be made for the geometries sampled by the GMM method, but with the new geometries concentrated more in the 70-90~kcal/mol. When attempting to predict the nitrogen inversion barrier in ammonia, only the ensemble-based method improved after two rounds of adversarial sampling and produced accurate predictions (Figure~\ref{fig:ammonia}e). However, after two rounds of adversarial sampling, the fraction of stable MD trajectories produced by the ensemble-based model, MVE, and evidential regression improved by roughly 30-40\% points (Figure~\ref{fig:ammonia}g). GMM, on the other hand, achieved an improvement of 52\% points, indicating that the quality of uncertainty estimates improved robustness of the NNIP outside of the training domain more than the other methods.

\subsection{Silica Glass}

\begin{figure}[ht]
    \centering
    \includegraphics[width=\linewidth]{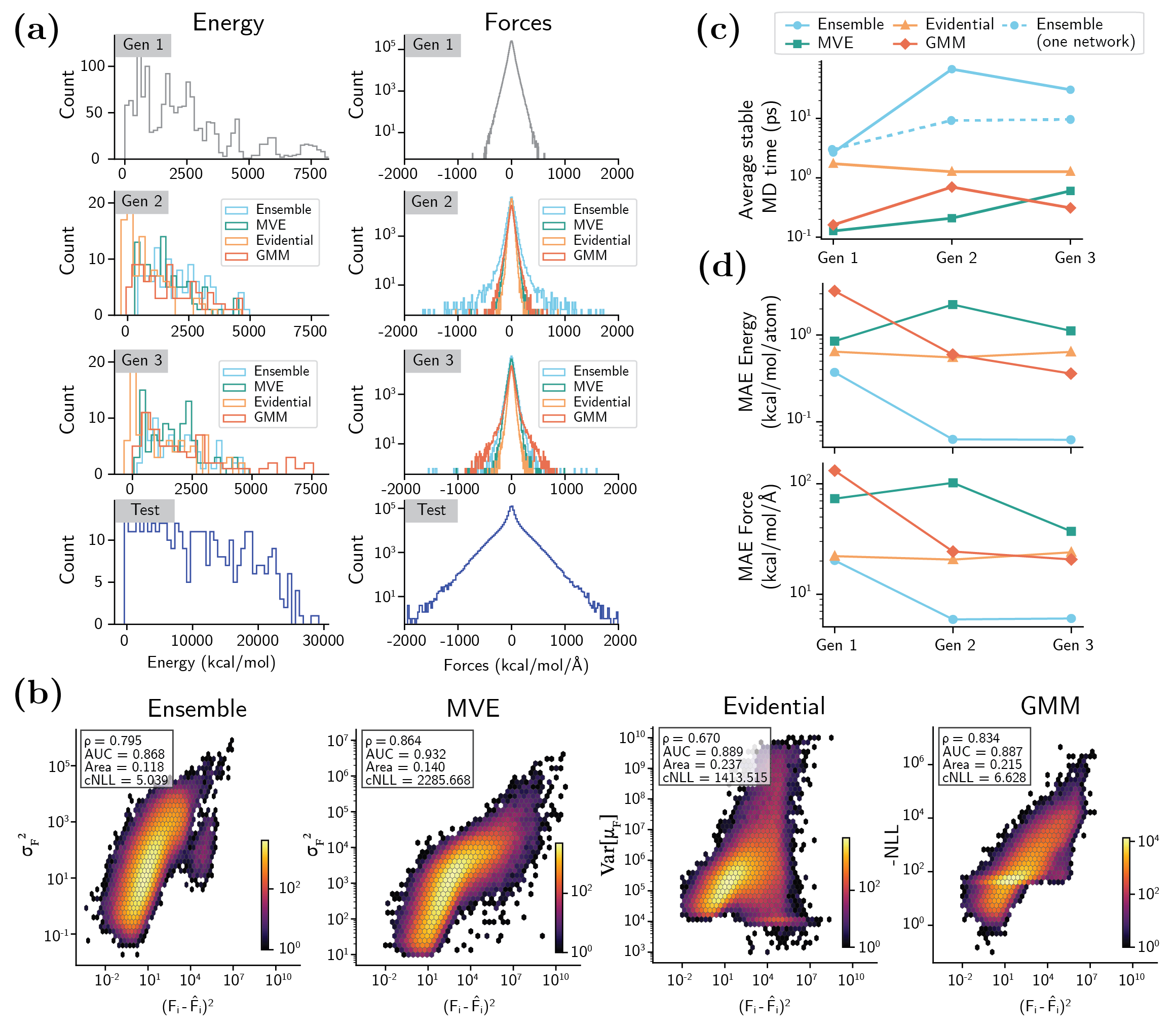}
    \caption{
    \textbf{(a)} Distributions of energy and forces in training and testing sets. In the first generation (first row), all models are trained on the same initial training set. In subsequent generations, NN models are trained on both initial training data and new sampled data points using their corresponding UQ methods. All models are finally evaluated on the test set, where the energy and force magnitudes are considerably larger than the train sets. Note that the x-axis scale of the test set energy distribution (last row first column) is different from the other energy distributions.
    \textbf{(b)} Hexbin plots showing the predicted uncertainties against squared errors of atomic forces from NNs trained only on initial training data. 
    \textbf{(c)} Average length of time in which no atomic dissociation or collision occurs in the MD trajectories. Average is taken from 10 trajectories.
    Item ``ensemble (one network)" in legend describes MD trajectories performed with predictions only from a single NN in the ensemble. 
    \textbf{(d)} Mean absolute error of predicted energy and forces by the NN models on the test set.
    }
    \label{fig:silica}
\end{figure}

Finally, we tested the UQ methods on silica glass, which has a long history for interatomic potential fitting\cite{VanBeest1990,Tangney2002,ttam,urata2021a,Pedone2022}. This is motivated by the fact that silica glass is made up of large bulk structures and the computational cost of performing quantum chemistry calculations on these bulk systems is very high. NNs, which have increasingly been used for interatomic potential fitting for vitreous silica\cite{urata2021b,Balyakin2020}, are unfortunately still unreliable outside of the training data domain. This calls for efficient sampling methods to generate small sets of additional data to improve high-error regions. These rely significantly on the quality of UQ methods, which should additionally be lightweight and fast. We trained individual NNs and an NN ensemble containing 4 networks on the same initial training data, and evaluated the errors and uncertainties from the respective UQ method on the test data set. Distributions of the initial training and testing data set are shown in Figure~\ref{fig:silica}a. The testing set contains structures with very high energies and forces which have not been trained or validated on any of the NNs, such that evaluation of the predicted uncertainties are guaranteed to be epistemic uncertainties outside of training data domain. Evaluated true errors and uncertainties, presented in Figure~\ref{fig:silica}b, for the UQ methods show that all methods are capable of ranking uncertainties relatively well with respect to the true errors, even when the evaluated metrics deliver inconsistent results as to which model is the best. Disputably, evidential regression seems to perform the worse in uncertainty ranking for this data set considering that the evaluated metrics, the order of magnitude of predicted uncertainties, and the distribution of uncertainty-error are less favorable. 

We then performed adversarial sampling using each of the UQ methods and obtained new data points which are evaluated using DFT and incorporated into re-training of the NNs in subsequent generations. Note that for NNs in each UQ method, the training data thus becomes different starting in generation 2 since the new data points are sampled using different UQ methods (Figure~\ref{fig:silica}a). In generation 2, energies of the adversarial samples for the methods have similar distributions with the exception of evidential regression, which interestingly did not drive the adversarial samples to very high energies but to energies lower than 0~kcal/mol (the reference energy). The reference energy is chosen as the minimum energy of the initial training data, which consists only of structures extracted from MD at 300~K or higher (refer to Figure~S1), and evidential regression instead sampled structures with energies lower than those included in the training data. Ensemble-based method, on the other hand, sampled structures that are within 0-5000 kcal/mol energy ranges, but contain high atomic forces. While this is arguably the constraint constructed for adversarial sampling (Equation \ref{eq:thermo_likelihood}), the effect of this constraint is most apparent in the ensemble-based method, where these new data points significantly increased the robustness of the NNIP. In the third generation, GMM sampled new geometries with both energies and atomic forces much higher than the other methods. 

After training of NNs in each generation, 10 MD simulations are performed using the NNs. The lengths of time where the simulations are stable are recorded in Figure~S15. Averages are plotted in Figure~\ref{fig:silica}c. The figures show that the simulations for generation~1 behaved unphysically early on (training was conducted only on the initial data). This is potentially due to prediction of false minima from the NN potentials that produce unphysical structures during the simulations\cite{Morrow2022HowPotentials}. Additionally, the length of time of stable MD trajectories differ despite the shared training data and model architecture. This is consistent with the results on the ammonia data set, where the robustness of NNIPs is directly influenced by the accuracy of the predictions, represented as mean absolute error of the energy and atomic forces (Figure~\ref{fig:silica}d and Figure~S14). In generation 2, the length of stable MD time for ensemble-based method increases greatly alongside the accuracy of prediction, potentially as a result of the inclusion of new adversarial data with very high forces. A similar trend can be observed for GMM, but at a smaller magnitude of improvement. However, in generation 3, there seems to be a reversal of improvement for both ensemble-based and GMM methods even though more data has been included and prediction errors on test set shown in Figure~\ref{fig:silica}d have decreased. This suggests that adding more data does not always improve robustness of NNs and that quality of new data is more crucial\cite{Koh2017UnderstandingFunctions}. MVE shows steady improvement but all MD simulations were unable to exceed 1~ps. In contrast, evidential regression did not show improvement with inclusion of more structures that were obtained through adversarial sampling (Figure~S16). Since the gradients of the different UQ measures could be different, we set both the Boltzmann temperatures and learning rates in adversarial sampling to be the same for all methods, but increased the number of iterations such that the uncertainties could be maximized. For this data set, we can see that ensemble-based method not only produces higher stability simulations, it also samples higher quality uncertain data in a way that cannot be quantified using the evaluation metrics. 

\subsection{General Remarks}
In general, there seems to be a trade-off between the prediction accuracy, confidence of UQ, and computational costs among the methods. For ensemble-based methods, the prediction accuracy and UQ performance demonstrate that it is more reliable than the other methods. However, the time and memory usage incurred for training the model scales with the number of NNs used in the ensemble and can be quite infeasible for extremely large data sets. MVE, on the other hand, performs relatively well in ranking uncertainty with respect to the true error and is computationally cheaper. The downside to this method is the lower prediction accuracy achieved as compared to other methods even with the same training parameters and conditions. Additionally, the predicted uncertainty optimized using the loss function during training (Eq.~S3) intrinsically describes only the aleatoric uncertainty. For training data generated via \textit{ab initio} methods, the data set is relatively noise-less and technically does not have meaningful aleatoric uncertainty, meaning MVE has somehow co-opted epistemic uncertainty. In the case of evidential regression, the method performs remarkably well in terms of uncertainty estimation but achieves a slightly lower prediction accuracy than the ensemble-based method. Nevertheless, it is important to note that training of the evidential regression method often incurs numerical instability due to optimization of the loss function (Eq.~S6) that drives the parameters ($\nu$, $\alpha - 1$, and $\beta$) close to zero. As for GMM, the model is easy to implement, computationally cheap, and can be used with any NN without compromising the achieved prediction accuracy. The downsides to this method, however, are the need for manual configuration of the number of Gaussians for each data set (Figures~S3 and~S4) and the reliance on the accuracy of Gaussian mixture model fitting. 

\section{Conclusions}

The performance of the uncertainty quantification (UQ) methods vary broadly between data sets based and evaluation metrics.
The ensemble-based UQ method performs consistently well across all data sets and metrics in this work. In addition, due to the averaging effect, ensemble predictions typically show lower prediction error albeit at a higher cost. However, other work has shown that ensemble-based method can only be used to address epistemic uncertainty stemming from parametric uncertainty (model variance), and is less reliable in identifying data points outside of the training domain\cite{Heid2023}. Contrary to these findings, our study reveals that the ensemble-based method is capable of identifying data points outside of the training domain when performing adversarial sampling.

Mean-variance estimation (MVE), on the other hand, has shown inconsistent performance across the data sets. We infer that MVE is less capable in systematically predicting high uncertainty for underrepresented regions, but is better at identifying regions within the training domain that correspond to high errors.
Deep evidential regression UQ has been shown to be decent but does not outperform the ensemble-based method and tends to be numerically unstable during parameter optimization.
The Gaussian mixture model (GMM) approach shows good performance in UQ, especially for out-of-domain frames, and is more versatile since it is applicable, post-training, to any model architecture, although not as well in the interpolative regime where MVE seemed to excel.

Nevertheless, for applications in neural network interatomic potentials (NNIPs), generalizability plays a very important role to ensure robustness of the atomistic simulations. A model that seemingly estimates uncertainty well and achieves good prediction accuracy but fails to achieve stable production simulations from scratch or through active learning, is less helpful. The unsolved challenge, thus is identifying a model that is able to rank uncertainty well, while at the same time scales practically, has high prediction accuracy, and produces stable atomistic simulations. UQ and robustness in NNIPs may pose a unique challenge because forces are a derivative function, which is easily accessible from the energy at training time, but the relationship between force variance and energy variance is not handled well by any of the methods explored. 

Albeit cheaper than ensemble-based method, single deterministic NNs showed lower generalizability, and thus failed to exceed the performance of the ensemble-based method, especially in the active learning loops, since ensemble learning lowers the parametric uncertainty. This is especially true for MVE since it has been shown that joint optimization of the mean and variance estimators lowers the stability of prediction\cite{Seitzer2022}. When a single NN architecture shows higher generalizability than that of a NN ensemble, then usage of UQ methods such as deep evidential regression or GMM for the single NNs would be justified. 

More broadly, the lack of a one-size-fits-all solution, the high cost of the more robust ensemble-based method, and the disappointing performance of elsewhere-promising novel evidential approaches confirm that UQ in NNIPs is an ongoing challenge for method development in AI for science.

\section{Acknowledgements}
This work was supported by funding from the AGC, Inc. 

\section{Declarations}
\begin{itemize}

    \item \textbf{Competing interest} The authors declare no competing interests.
    
    \item \textbf{Data availability} The molecular dynamic trajectories generated in the current study have been deposited in the Materials Cloud Archive under accession code \url{https://doi.org/10.24435/materialscloud:55-sd}\cite{Tan2023data}.
    
    \item \textbf{Code Availability} The code used to produce the results in this paper is available at \url{https://github.com/learningmatter-mit/UQ_singleNN.git} under the MIT license. 
    
    \item \textbf{Author contributions} R.G.-B and A.R.T. conceived the project. A.R.T. designed and conducted the experiments, performed data analysis and wrote the manuscript. S.U. performed molecular dynamics simulations and DFT calculations on the initial silica training data set. S.G. and J.C.B.D. provided fruitful advice and discussions. J.C.B.D. contributed to content organization and manuscript review. R.G.-B. supervised the research, secured the funding and contributed to manuscript writing.

\end{itemize}

\newpage

\bibliography{references}



\section*{Supplementary material}

**Supporting information for**

**Single-model uncertainty quantification in neural network potentials does not consistently outperform model ensembles**

Aik Rui Tan,[1] Shingo Urata,[2] Samuel Goldman,[3] Johannes C. B. Dietschreit,[1] and Rafael Gómez-Bombarelli[1, a)]

[1)] *Department of Materials Science and Engineering, Massachusetts Institute of Technology*

[2)] *Innovative Technology Laboratories, AGC Inc.*

[3)] *Computational and Systems Biology, Massachusetts Institute of Technology*

(Dated: 2 May 2023)

---

[a)]Electronic mail: rafagb@mit.edu

## I. NEURAL NETWORK INTERATOMIC POTENTIALS (NNIPS)

We used Polarizable atom interaction Neural Network (PaiNN), an equivariant message-passing NN to learn interatomic potentials of molecules and materials. The NNIP architecture was the same for all systems, we used 3 convolutional layers with a hidden dimension size of 128 each, 20 learnable Gaussians, a cutoff of 5.0 Å and shifted softplus activation function. The 128-dimensional feature vectors were then passed through 2 dense layers of size 64 before being summed to yield a final output value. For a specific system, all models were trained on different splits of the data set using the Adam optimizer with an initial learning rate of $10^{-4}$. A scheduler was used to reduce the learning rate by a factor of 2 if the error on the validation set stagnates for 30 epochs.

For the rMD17 data set, all models were trained and validated on 800 and 200 data points (molecular conformations) in each fold/split for 500 epochs with a batch size of 64. The models were then tested on 1000 data points. 5-fold cross validation was performed for each molecule and UQ method, where the indices for the train and test set for each fold was provided by Christensen et al. For the ammonia data set, first generation models were trained on 62 geometries and validated on 16 geometries to assess the models' ability to bootstrap on few data points. The test set contains 200 geometries with energies ranging from 0 to 100 kcal/mol. The models were trained for 500 epochs with a batch size of 64. For the silica data set, the models were trained on a data set containing 1590 geometries (train: validation at a ratio 80: 20) and tested on 402 structures from an external test set with energies ranging from 0 - 30,000 kcal/mol (more details in the data set section). They were trained for 1000 epochs with a batch size of 3.

### A. Ensemble-based method

5 independent NNs are used for rMD17 and ammonia data sets, whereas 3 independent NNs are used for the silica glass data set. During training of the NNs in the ensembles, the loss $\mathcal{L}$ is then computed as the mean absolute error of the predicted and target values of a batch size $N$

$$\mathcal{L} = \frac{1}{M}\frac{1}{N}\sum_{m}^{M}\left[\sum_{i}^{N}\left(w_E|E_{i,m} - \hat{E}_{i,m}| + w_F|\mathbf{F}_{i,m} - \hat{\mathbf{F}}_{i,m}|\right)\right] \quad (1)$$

where $w_E$ and $w_F$ are the weights of the energy and force trade-offs during training. In all models, the weights were set to 0.1 and 1.0, respectively.

## B. Mean-variance estimation (MVE)

NNs in the MVE method is trained using a negative log-likelihood (NLL) loss function of the normal distribution instead of a regular mean absolute error (MAE). In our work, the NLL loss function was applied only on the force prediction while a regular mean absolute error (MAE) was applied on the energy outputs. This increases the prediction accuracy of the NN. Similar to the ensemble method, the uncertainty estimates are described only by the variances of the atomic forces.

$$\text{MAE}_E = \frac{1}{N} \sum_i^N |E_i - \mu_{E_i}| \tag{2}$$

$$\text{NLL}_F = \frac{1}{2N} \sum_i^N \log(2\pi\sigma_{F_i}^2) + \frac{(\mathbf{F}_i - \mu_{F_i})^2}{2\sigma_{F_i}^2} \tag{3}$$

$$\mathcal{L} = w_E \text{MAE}_E + w_F \text{NLL}_F \tag{4}$$

where $w_E$ and $w_F$ are set to 0.1 and 1.0, respectively.

## C. Deep evidential regression

For training of the deep evidential NNs, the loss function used is computed as

$$\text{MAE}_E = \frac{1}{N} \sum_i^N |E_i - \gamma_{E_i}| \tag{5}$$

$$\text{NLL}_F = \frac{1}{N} \sum_i^N \frac{1}{2} \log\left(\frac{\pi}{\nu}\right) - \alpha \log\left[2\beta(1+\nu)\right] \tag{6}$$

$$+ \left(\alpha + \frac{1}{2}\right) \log\left\{(\mathbf{F}_i - \gamma_{F_i})^2 \nu + [2\beta(1+\nu)]\right\} + \log\left(\frac{\Gamma(\alpha)}{\Gamma(\alpha + \frac{1}{2})}\right) \tag{7}$$

$$+ \lambda |\mathbf{F}_i - \gamma_{F_i}| \cdot (2\nu + \alpha) \tag{8}$$

$$\mathcal{L} = w_E \text{MAE}_E + w_F \text{NLL}_F \tag{9}$$

where $\lambda$, which we set to 0.2, controls the trade-off between minimizing evidence on errors (i.e. lowering $\nu$ and $\alpha$ for incorrect predictions) and maximizing model fit. Again, $w_E$ and $w_F$ are set to 0.1 and 1.0, respectively.

### D. Gaussian mixture model (GMM)

In GMM method, a NN is trained with a regular mean absolute error (MAE) function as shown in Equation 1 where $M = 1$, $w_E = 0.1$, and $w_F = 1.0$

## II. DATASET

### A. Silica

The silica glass ($SiO_2$) training data set consists of 1590 structures, sampled from multiple long molecular dynamics trajectories subjected to different conditions including temperature equilibration, cooling, uniaxial tension, uniaxial compression, biaxial tension, biaxial compression, and biaxial shear at different rates using the force-matching potential described described in Urata et al.[1]. Each amorphous silica structure consists of 699 atoms (233 Si and 466 O atoms). In simulations of the cooling condition, 5 silica structures were cooled from 2300 K to 300 K at a rate of 1 K/ps and frames were saved every 25 ps. As for the temperature equilibration simulations, the amorphous silica structures were equilibrated at 300 K, 500 K, 1000 K, 1500 K, 2000 K, 2500 K, 3000 K, and 3500 K for 11 ns, and the frames were saved every 100 ps from the last 10 ns of the simulations. Note that the cooling simulations are separate from the temperature equilibration simulations. In the uniaxial deformation condition, we did two different simulations, one each for uniaxial tension and uniaxial compression. Specifically, the silica systems were subjected to a constant strain rate of $\pm 10^{-4}$ ps$^{-1}$ (+ indicating tension and − indicating compression), reaching a final strain magnitude of $\pm 0.30$ in one direction and the frames were sampled every 100 ps corresponding to a change in strain of $\pm 0.1$. In the biaxial tension and compression cases (separate simulations), systems are subjected to a constant strain rate of $\pm 0.0001$ ps$^{-1}$ in the x-direction up to $\pm 0.30$ final strain magnitude. In the y-direction, strain rates of $\pm 2.5 \times 10^{-5}$ ps$^{-1}$,

---

[1] Urata, S., Nakamura, N., Aiba, K., Tada, T., Hosono, H. **Mater. Des.**, 197,109210. https://doi.org/10.1016/j.matdes.2020.109210 (2021).

$\pm 5.0 \times 10^{-5}$ ps$^{-1}$, $\pm 7.5 \times 10^{-5}$ ps$^{-1}$, and $\pm 10^{-4}$ ps$^{-1}$ were applied, reaching final strain magnitudes of $\pm 0.075$, $\pm 0.015$, $\pm 0.0225$, and $\pm 0.3$, respectively. In the biaxial shear simulations, tensile stresses are applied in the x-direction to the systems at a strain rate of $\pm 10^{-4}$ ps$^{-1}$ up to a maximum strain of -0.3. In the y-direction, compressive stresses are applied at strain rates of $2.5 \times 10^{-5}$ ps$^{-1}$, $5.0 \times 10^{-5}$ ps$^{-1}$, $7.5 \times 10^{-5}$ ps$^{-1}$, and $10^{-4}$ ps$^{-1}$ up to maximum strains of 0.075, 0.015, 0.0225, and 0.3, respectively (See Table S1 for tabulated details of the simulations and Figure S1 for the data distributions). All frames were sampled every 100 ps. All MD simulations were done using the LAMMPS package using temperature and pressure controls of the Nosé-Hoover thermostat and barostat. Additionally, an independent test set was created from adversarial sampling during trial runs in this work. 402 structures with energies between 0 and 30,000 kcal/mol (43 kcal/mol/atom) were picked to form the test set. The structures were binned and deduplicated according to their energies and 2 structures with the highest atomic forces were chosen from each bin. No structures appear in both the train and test set.

In total, 1590 silica structures were sampled from the molecular dynamic trajectories to form the train set and 402 were selected from adversarial sampling to form the test set. DFT calculations were performed on the structures using the Vienna Ab-initio Simulation Package (VASP) within the projector-augmented wave (PAW) method. The Perdew-Burke-Ernzerhof (PBE) functional within the generalized gradient approximation (GGA) was adopted as the exchange-correlation functional. The kinetic energy cutoff for plane waves was restricted to 600 eV. The threshold for the energy convergence within a self-consistent field (SCF) cycle was set to $10^{-5}$ eV.

| Simulation Type | Name | Temperature (K) | Heating / cooling rate (K/ps) | Strain rate in x direction ($10^{-4}$/ps) | Strain rate in y direction ($10^{-4}$/ps) | Num frame |
|---|---|---|---|---|---|---|
| Temperature equilibration | 300 K | 300 | - | - | - | 100 |
| | 500 K | 500 | - | - | - | 100 |
| | 1000 K | 1000 | - | - | - | 100 |
| | 1500 K | 1500 | - | - | - | 100 |
| | 2000 K | 2000 | - | - | - | 100 |
| | 2500 K | 2500 | - | - | - | 99 |
| | 3000 K | 3000 | - | - | - | 98 |
| | 3500 K | 3500 | - | - | - | 92 |
| Cooling | cooling | - | 1.00 | - | - | 410 |
| Uniaxial deformation | uniaxial_compress | 300 | - | -1.00 | - | 30 |
| | uniaxial_stretch | 300 | - | 1.00 | - | 30 |
| Biaxial compression | biaxial_m1-m0.25 | 300 | - | -1.00 | -0.25 | 30 |
| | biaxial_m1-m0.5 | 300 | - | -1.00 | -0.50 | 30 |
| | biaxial_m1-m0.75 | 300 | - | -1.00 | -0.75 | 30 |
| | biaxial_m1-m1 | 300 | - | -1.00 | -1.00 | 30 |
| Biaxial stretch | biaxial_1-0.25 | 300 | - | 1.00 | 0.25 | 23 |
| | biaxial_1-0.5 | 300 | - | 1.00 | 0.50 | 26 |
| | biaxial_1-0.75 | 300 | - | 1.00 | 0.75 | 23 |
| | biaxial_1-1 | 300 | - | 1.00 | 1.00 | 22 |
| Biaxial shear | biaxial_1-m0.25 | 300 | - | 1.00 | -0.25 | 28 |
| | biaxial_1-m0.5 | 300 | - | 1.00 | -0.50 | 29 |
| | biaxial_1-m0.75 | 300 | - | 1.00 | -0.75 | 30 |
| | biaxial_1-m1 | 300 | - | 1.00 | -1.00 | 30 |
| | | | | | | 1590 |

Table S1. Table containing breakdown of the initial training dataset of silica.

### III. ADVERSARIAL SAMPLING

In the adversarial sampling strategy, a perturbation, $\delta$ is applied to randomly chosen structures, $x$ from the initial training data. The perturbation is then iteratively updated using gradient ascent technique to maximize the uncertainty of the perturbed structures, $U(x_\delta)$, as defined in Eq. 15 in the main paper. For our purposes, the perturbation update was optimized for 80 iterations at a normalized temperature $kT$ of 0.7 kcal/mol/atom. The learning rates were set to $5 \times 10^{-5}$ and $3 \times 10^{-5}$ for ammonia and silica, respectively.

20 adversarial conformers were sampled at each generation (i.e., active learning loop) for ammonia while 100 adversarial conformers were sampled for silica. Finally, we set a cutoff of 100 kcal/mol for new adversarial samples for ammonia such that the geometries are not extremely distorted. For silica, the cutoff was set at 5000 kcal/mol (7.2 kcal/mol/atom) for generation 2 and increased to 10000 kcal/mol (14.3 kcal/mol/atom) at generation 3.

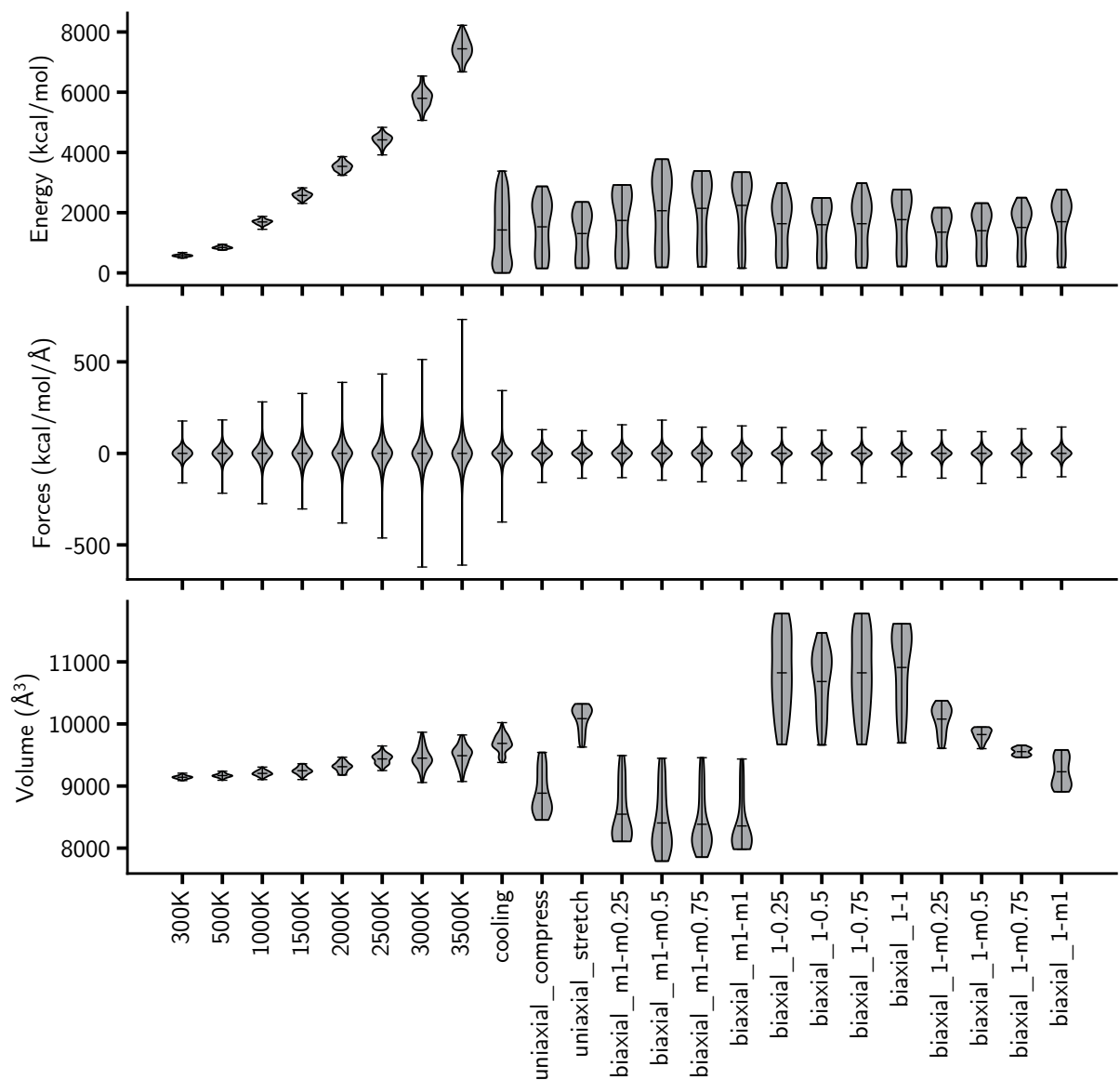

Figure S1. Violin plots showing energy, force and cell volume distributions of the initial silica dataset containing density functional theory (DFT) calculations of frames extracted from molecular dynamics (MD) trajectories performed under various conditions (See Table S1). All silica systems contain 699 atoms. In the biaxial tension cases, the first and second numbers after the underscore ("_") indicate the strain rates in the magnitude of $1 \times 10^{-4}$ ps$^{-1}$ in the x-direction and y-direction, respectively. "m" denotes minus, meaning that a compressive strain is applied.

## IV. ADDITIONAL RESULTS

### A. rMD17 dataset

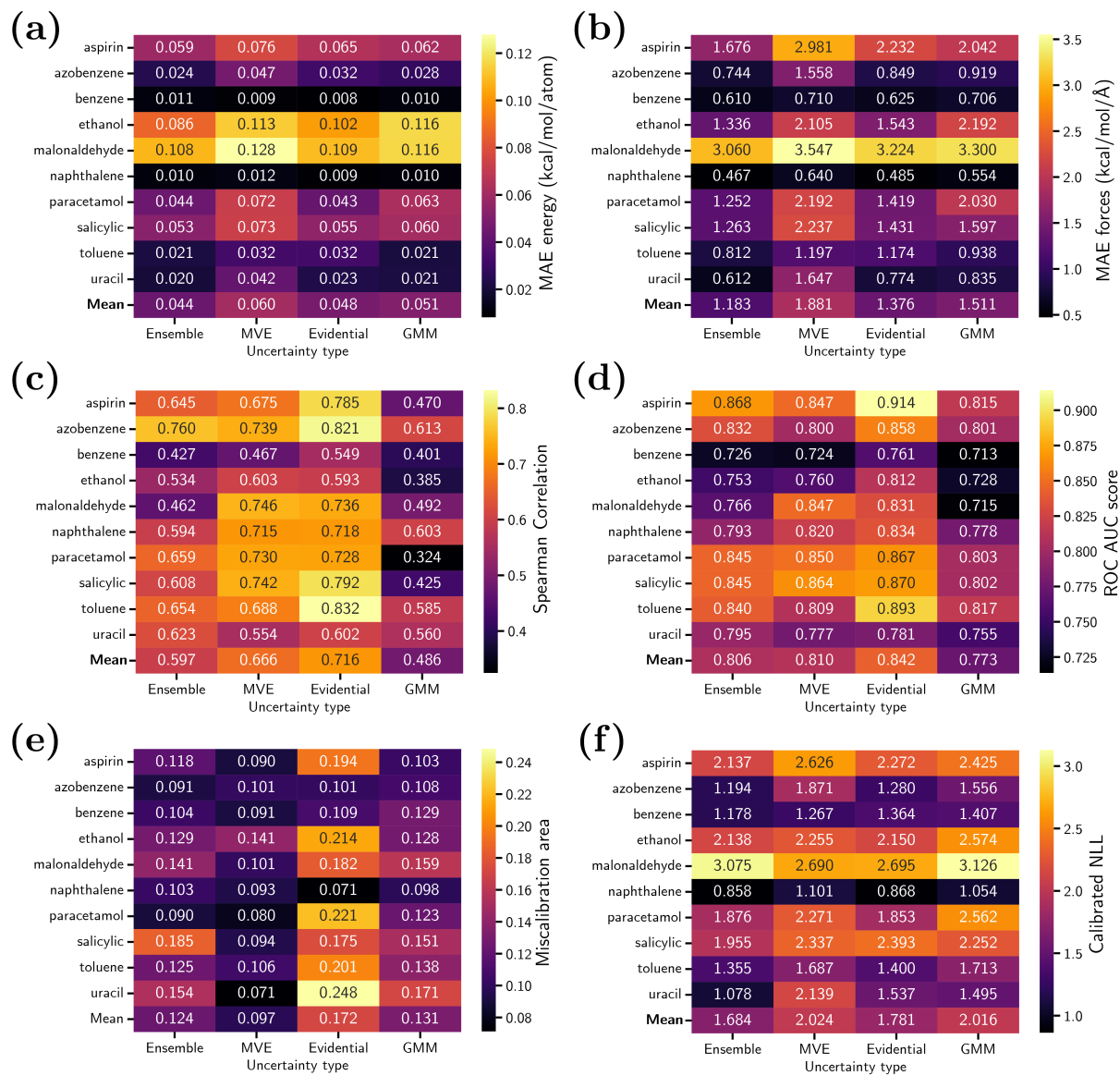

Figure S2. Heatmaps of prediction accuracies and evaluation metrics for uncertainty estimation on the test sets of rMD17 dataset for each considered uncertainty estimation methods. Each cell represents the average for all splits on each molecule. **(a)** Mean absolute error of energy in unit kcal/mol/atom. **(b)** Mean absolute error of atomic forces in unit kcal/mol/Å. **(c)** Spearman's rank correlation coefficient. **(d)** Area under the curve of receiver operating characteristics curve (ROC AUC score). **(e)** Miscalibration area. **(f)** Calibrated negative log-likelihood score.

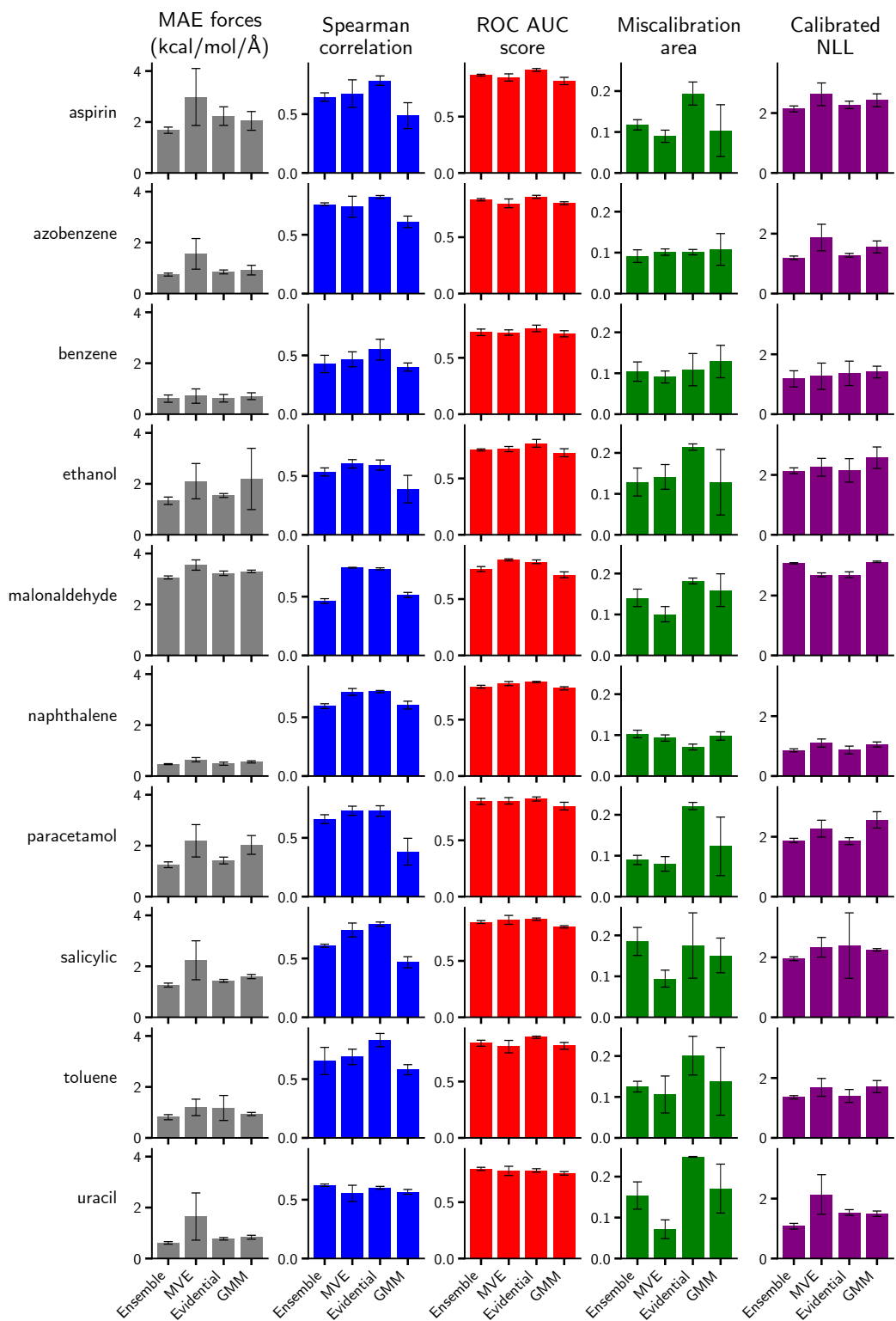

Figure S3. Bar charts showing easier comparison of the prediction accuracy (mean absolute error of atomic forces) and uncertainty estimation evaluation metrics on all molecules in the rMD17 dataset for each considered uncertainty estimation methods.

9

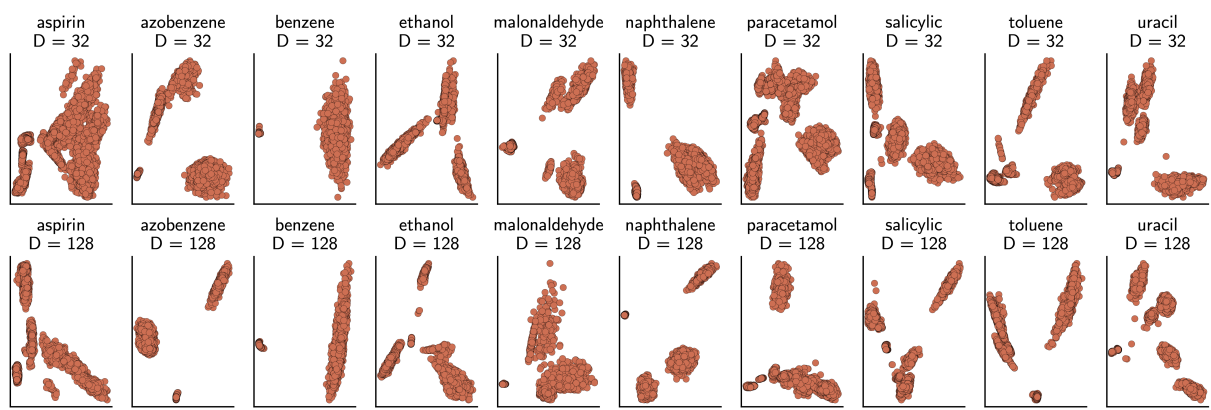

Figure S4. Principal component analysis (PCA) of the latent embeddings in neural networks (NNs) used for Gaussian mixture models (GMM) on each molecule in the rMD17 dataset (for the train set of one split). This is illustrated such that the number of clusters/components to fit the GMMs can be configured well.

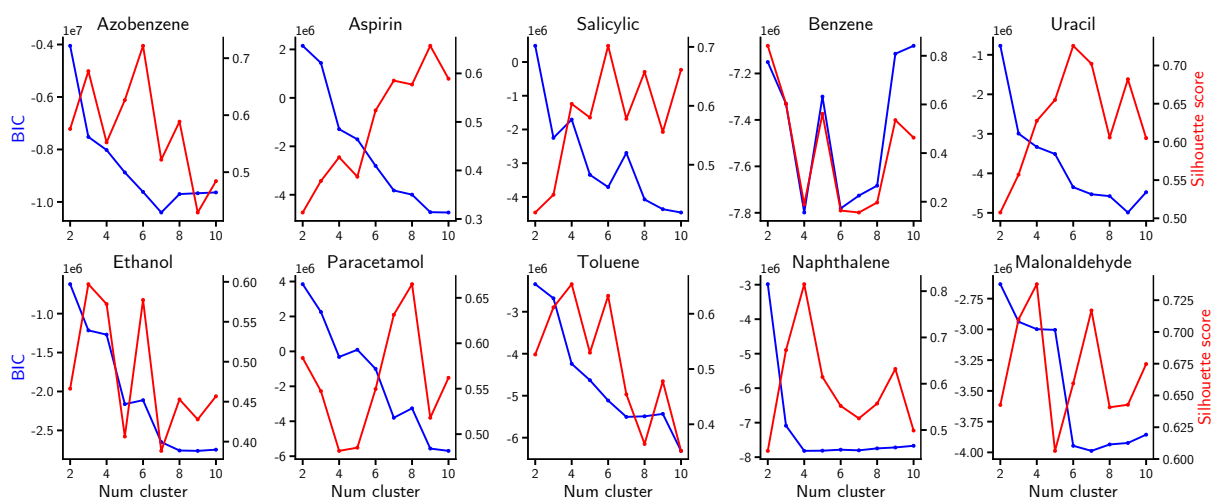

Figure S5. Plots showing Bayesian information criteria (BIC) and silhouette scores with respect to the number of clusters/components used for fitting latent embeddings in neural networks (NNs) used for GMM. The number of clusters can then be determined for optimal fitting of the GMMs by examining the tradeoff between the BIC and silhouette scores. A model that fit well would ideally have lower BIC and higher (closer to 1) silhouette scores.

10

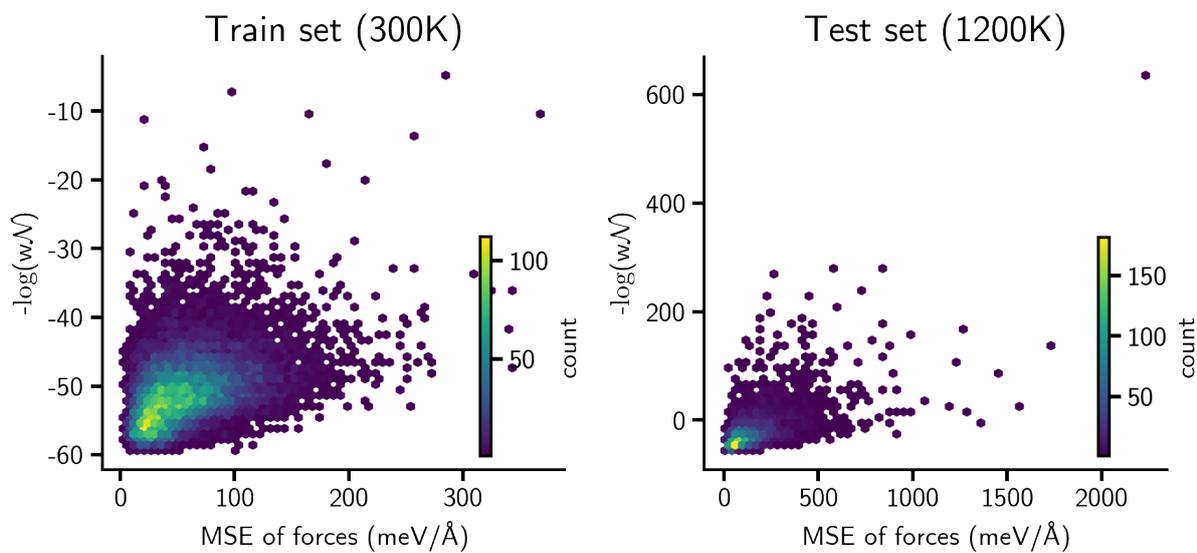

Figure S6. Replication of the 3BPA molecule results in the original paper proposing GMM[1] for uncertainty estimation scheme.

---

[2] A. Zhu, S. Batzner, A. Musaelian, and B. Kozinsky, Fast Uncertainty Estimates in Deep Learning Interatomic Potentials, arXiv:2211.09866, 1 (2022).

11

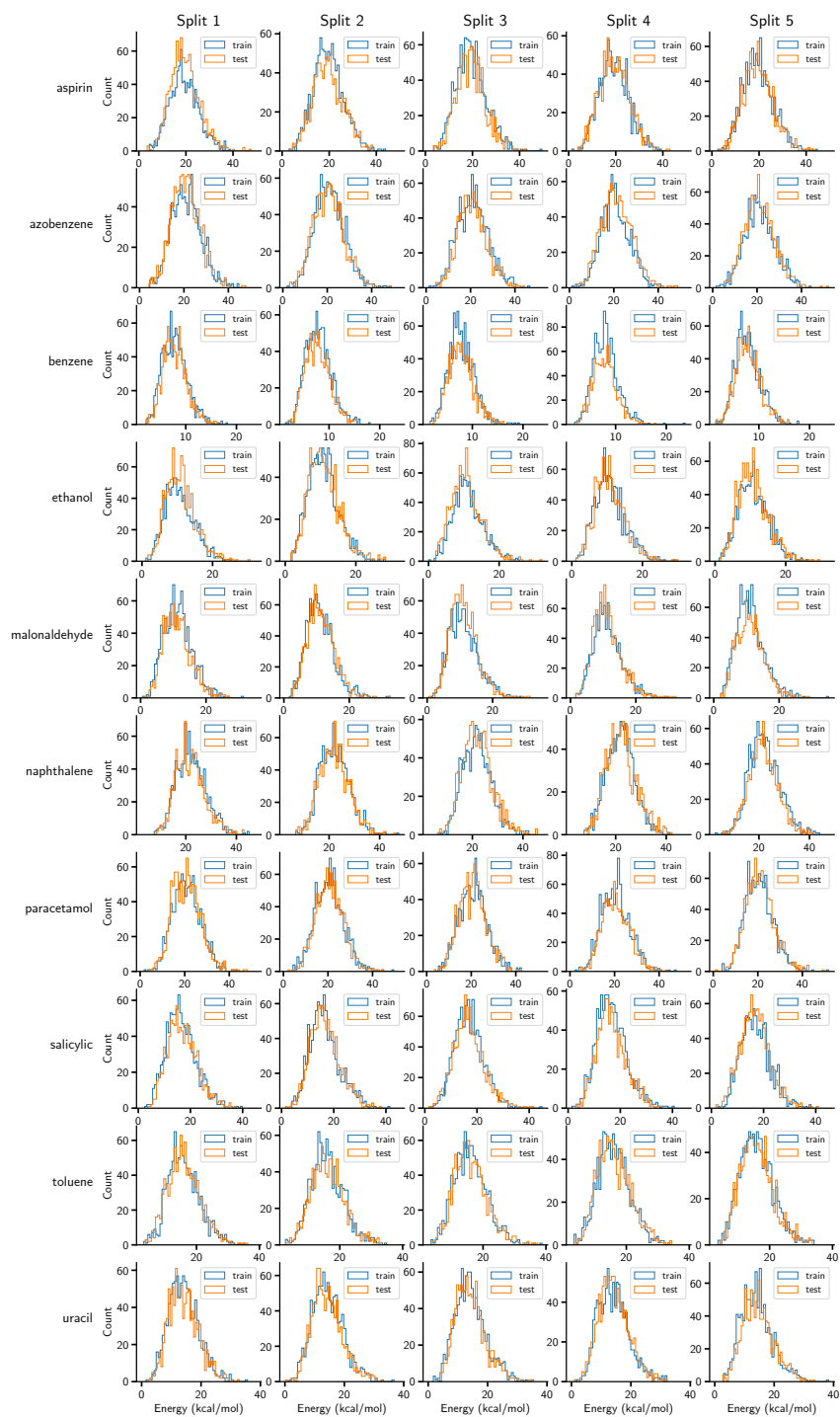

Figure S7. Energy distribution of train and test sets of the rMD17 dataset for all splits (across) of all molecules (down).

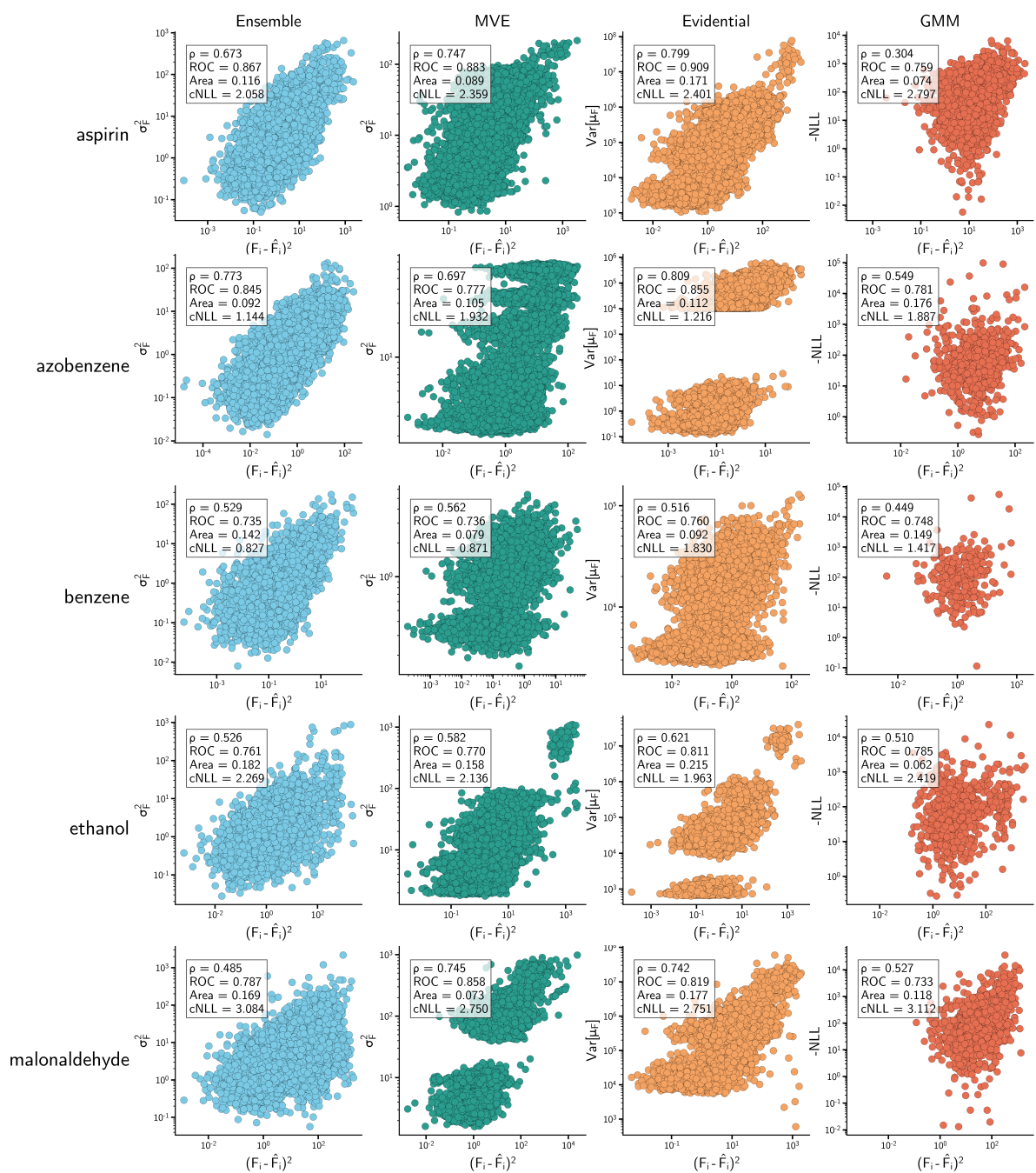

Figure S8. Scatter plots showing (predicted) uncertainties versus squared errors of atomic forces of the aspirin, azobenzene, benzene, ethanol, and malonaldehyde molecules in the rMD17 dataset for each considered uncertainty estimation method. Note that since the uncertainties (-NLL) for the GMM method contain negative values, all uncertainties are scaled by the minimum value to yield positive quantities for plotting of the log scale figure.

13

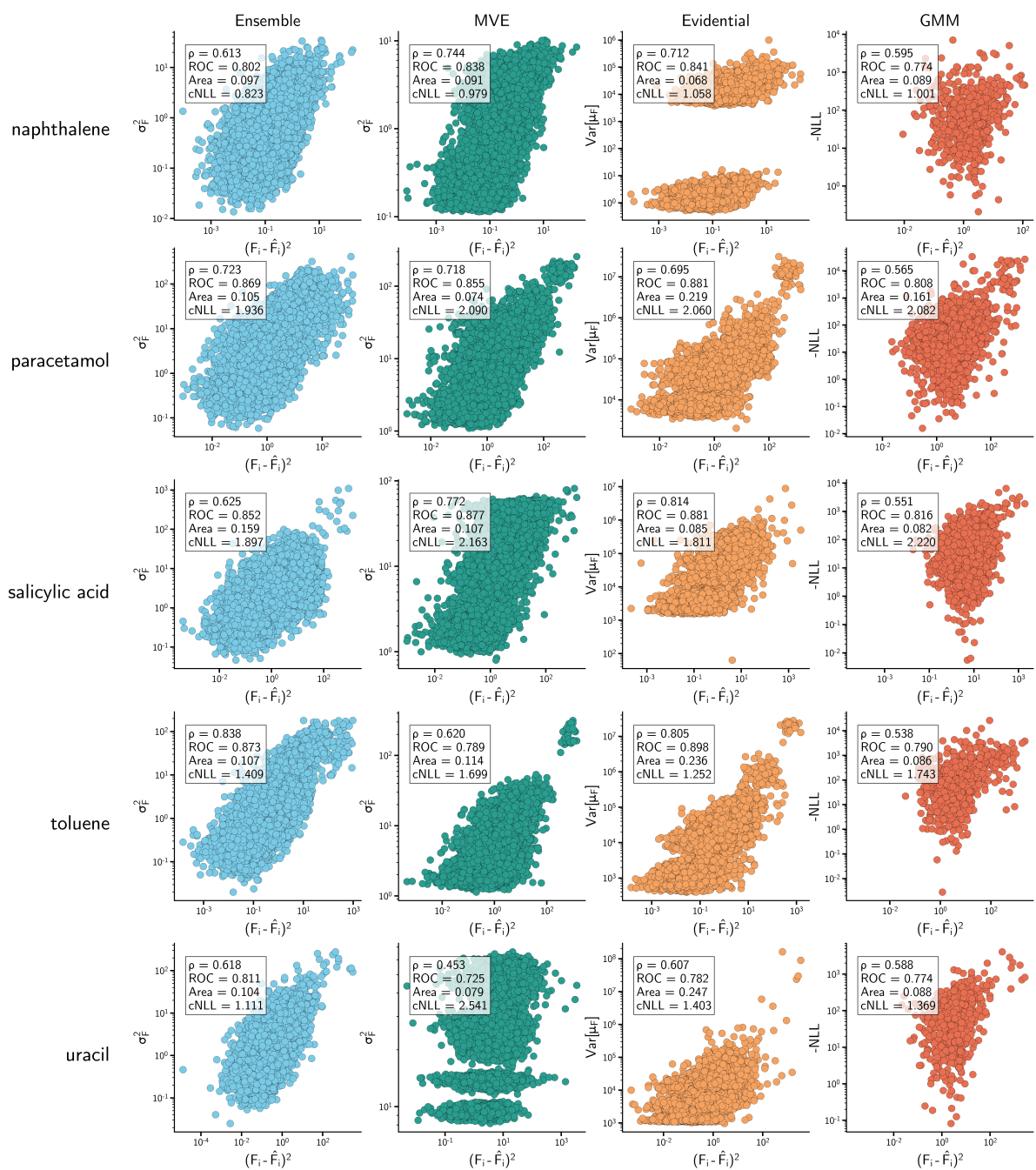

Figure S9. Scatter plots showing (predicted) uncertainties versus squared errors of atomic forces of the naphthalene, paracetamol, salicylic acid, toluene, and uracil molecules in the rMD17 dataset for each considered uncertainty estimation method. Note that since the uncertainties (-NLL) for the GMM method contain negative values, all uncertainties are scaled by the minimum value to yield positive quantities for plotting of the log scale figure.

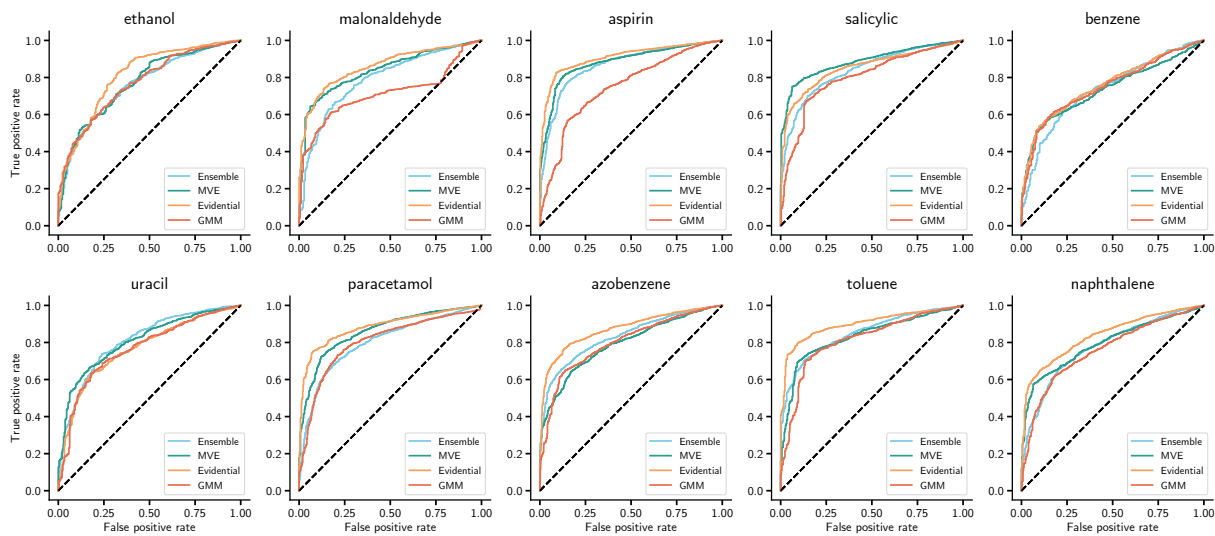

Figure S10. Receiver operating characteristics (ROC) curve of each considered uncertainty estimation method for all molecules in the rMD17 dataset. The threshold specified for the "target" (true error) is at 20th percentile of the distribution.

|  | 20th percentile | | | | 50th percentile | | | |
|---|---|---|---|---|---|---|---|---|
|  | Ensemble | MVE | Evidential | GMM | Ensemble | MVE | Evidential | GMM |
| aspirin | 0.853 | 0.840 | 0.921 | 0.783 | 0.815 | 0.834 | 0.887 | 0.733 |
| azobenzene | 0.853 | 0.826 | 0.890 | 0.807 | 0.883 | 0.873 | 0.915 | 0.821 |
| benzene | 0.700 | 0.716 | 0.764 | 0.702 | 0.702 | 0.733 | 0.777 | 0.698 |
| ethanol | 0.745 | 0.765 | 0.780 | 0.693 | 0.755 | 0.800 | 0.779 | 0.679 |
| malonaldehyde | 0.763 | 0.866 | 0.849 | 0.745 | 0.741 | 0.887 | 0.876 | 0.786 |
| naphthalene | 0.786 | 0.840 | 0.843 | 0.781 | 0.788 | 0.868 | 0.867 | 0.800 |
| paracetamol | 0.837 | 0.862 | 0.874 | 0.773 | 0.822 | 0.870 | 0.854 | 0.690 |
| salicylic | 0.835 | 0.884 | 0.904 | 0.794 | 0.811 | 0.876 | 0.901 | 0.738 |
| toluene | 0.842 | 0.821 | 0.924 | 0.821 | 0.808 | 0.839 | 0.912 | 0.784 |
| uracil | 0.793 | 0.771 | 0.787 | 0.758 | 0.805 | 0.779 | 0.802 | 0.779 |
| Mean | 0.801 | 0.819 | 0.854 | 0.766 | 0.793 | 0.836 | 0.857 | 0.751 |

Table S2. Area under the curve of receiver operating characteristics (ROC AUC) score of the considered uncertainty estimation methods in the rMD17 dataset using the 20th and 50th percentile to determine the binary cutoff in the true error (target) distribution.

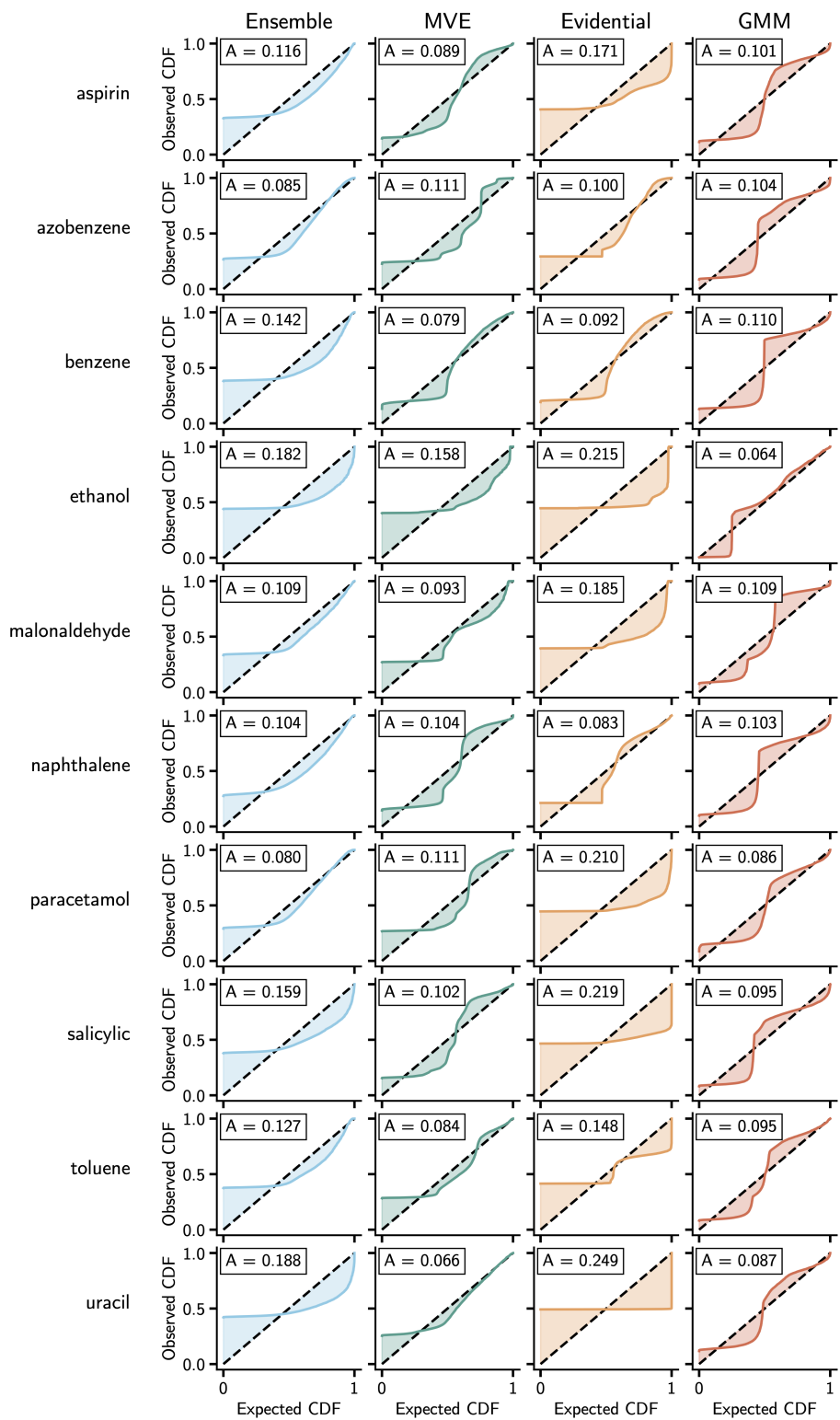

Figure S11. Calibration curves of each considered uncertainty estimation method (across) for all molecules in the rMD17 dataset. Boxes at the top left of each subfigure indicates the miscalibration area, which is the shaded area between the calibration curve and the ideal curve (dashed).

16

## B. Ammonia dataset

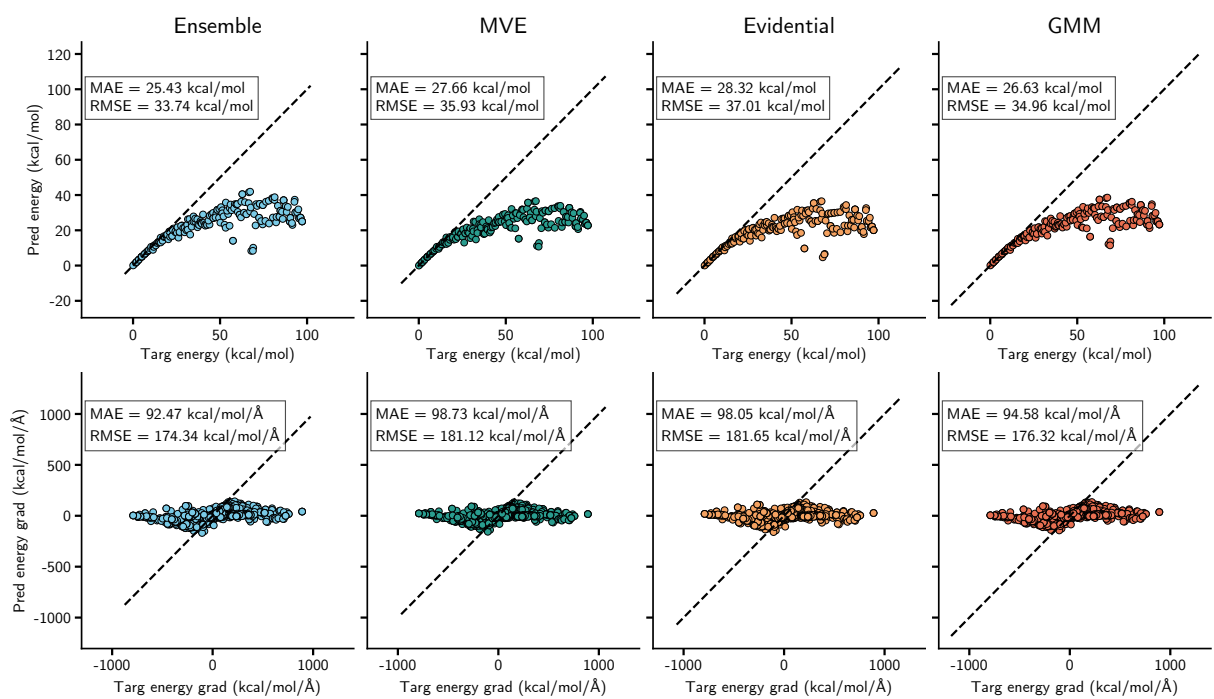

Figure S12. Parity plot showing predicted energy and forces on the test set of ammonia using the NN models trained in generation 1 (trained only on 78 geometries).

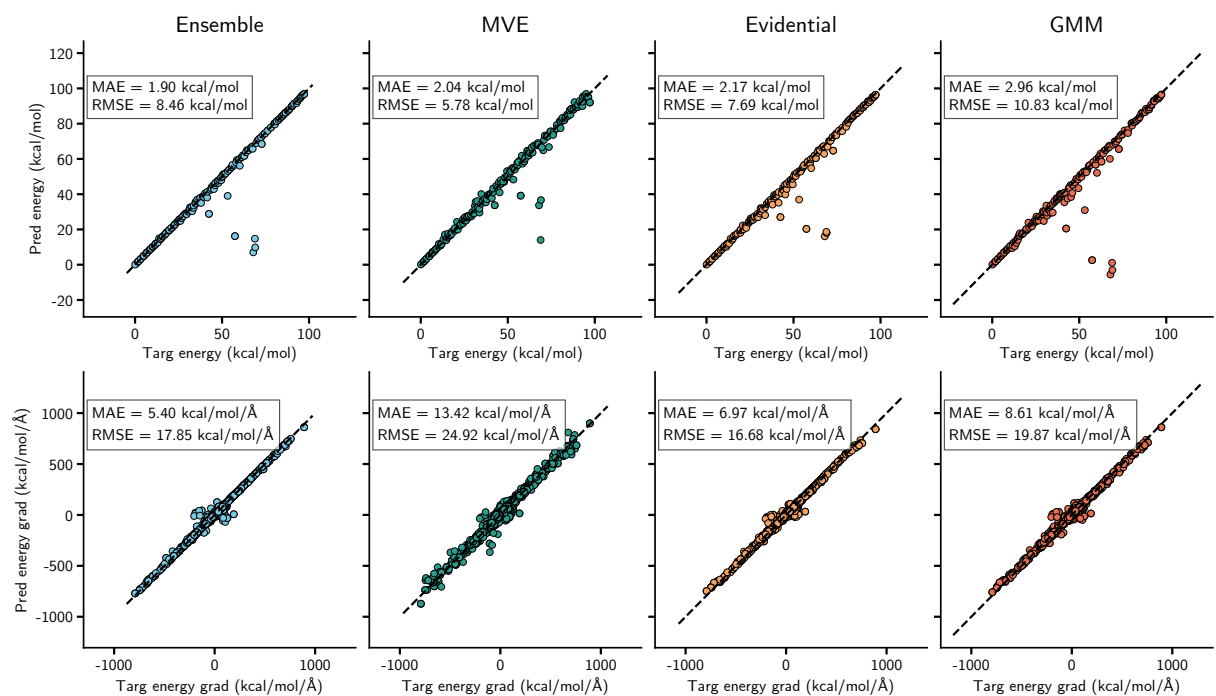

Figure S13. Parity plot showing predicted energy and forces on the test set of ammonia using the models trained in generation 3 (trained on 78 initial geometries and 40 adversarial samples).

## C. Silica dataset

**(a) Train**

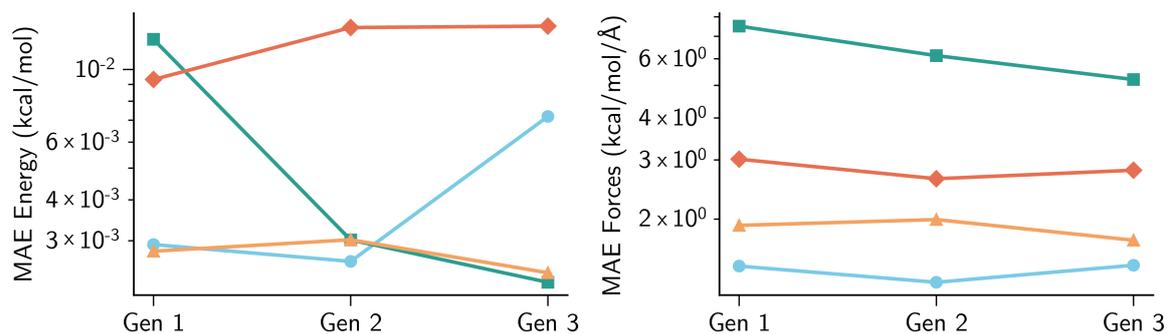

**(b) Validation**

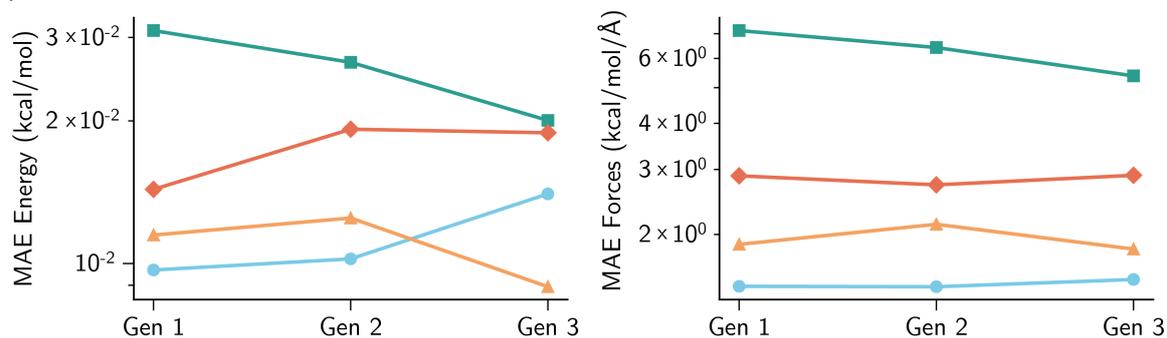

**(c) Test**

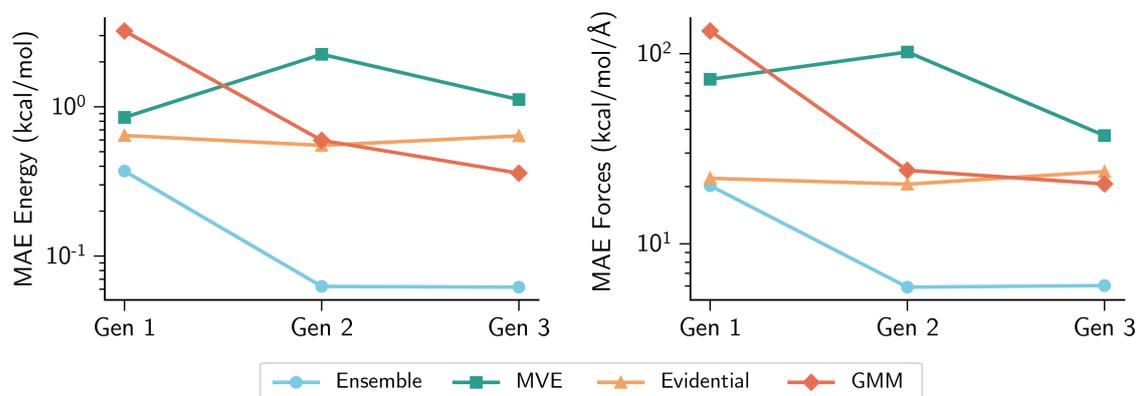

Figure S14. Mean absolute error of predicted energy (first column) and forces (second column) of NN models with respect to different generations on the **(a)** training set, **(b)** validation set, and **(c)** testing set of silica.

19

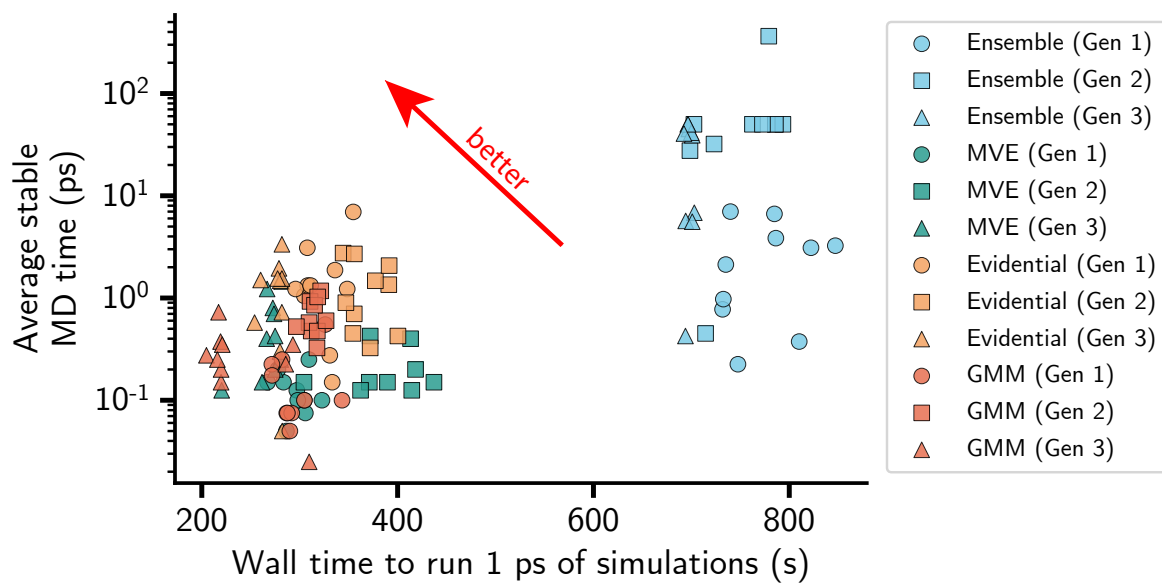

Figure S15. Scatter plots showing the length of stable MD trajectories against the wall time taken to run 1 ps of simulations for NN models in each considered uncertainty estimation methods in different generations. Every marker indicates a 1 ns-long NVT trajectory simulated at 2500K.

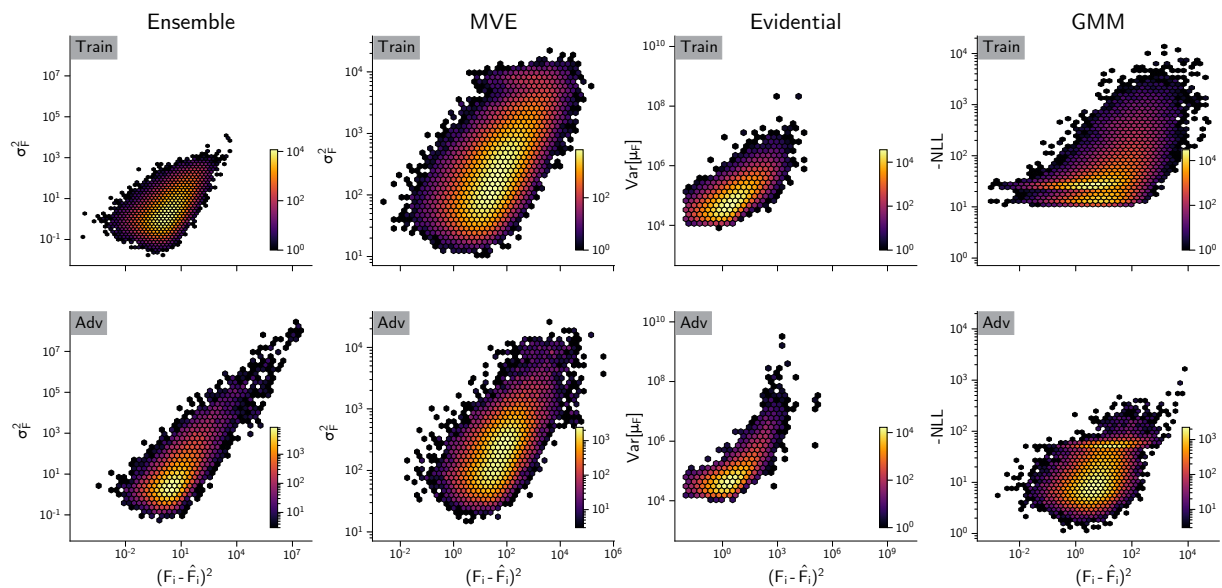

Figure S16. Hexbin plots showing the predicted uncertainties against true errors from NN models of the uncertainty estimation methods in generation 1. The top row presents the distribution of the train set and the bottom row shows distribution of adversarial samples generated by the respective uncertainty estimation methods.

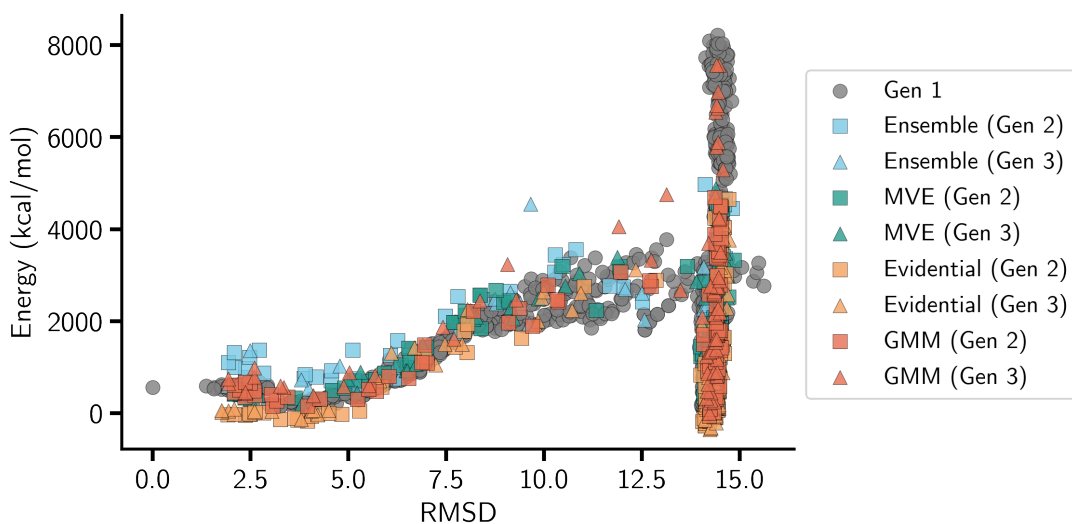

Figure S17. Energy against root mean squared deviation (RMSD) of silica structures in training data and new adversarial samples generated by the respective uncertainty estimation methods. RMSD of all silica structures are computed against a silica structure sampled at a 300 K temperature equilibration simulation in the training data.

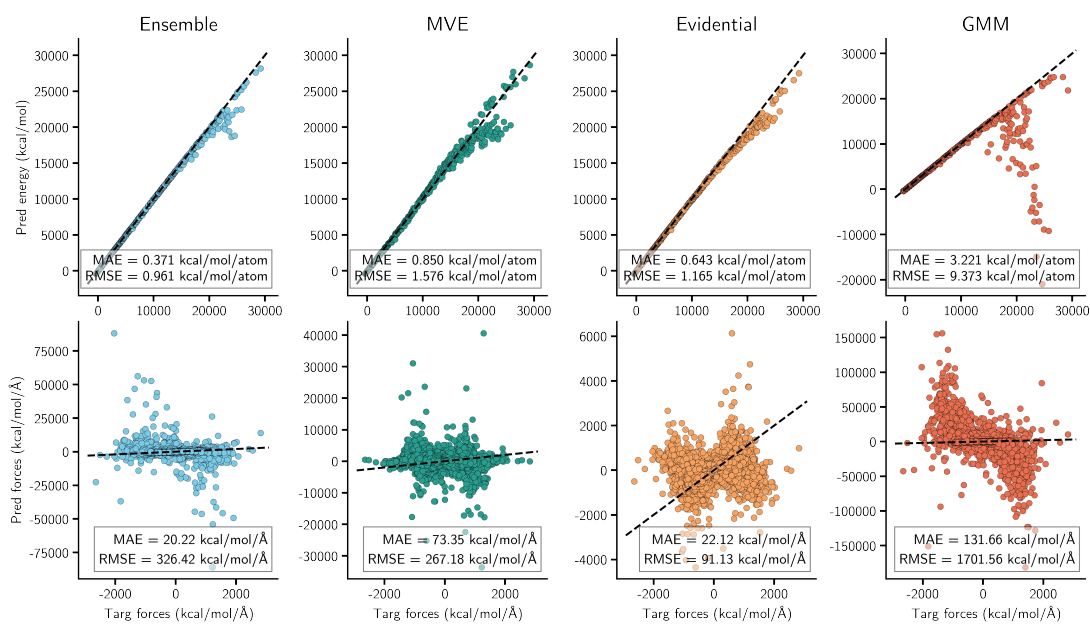

Figure S18. Parity plot showing predicted energy and forces on the test set of silica using the models trained in generation 1 (trained on 1590 initial training data).

21

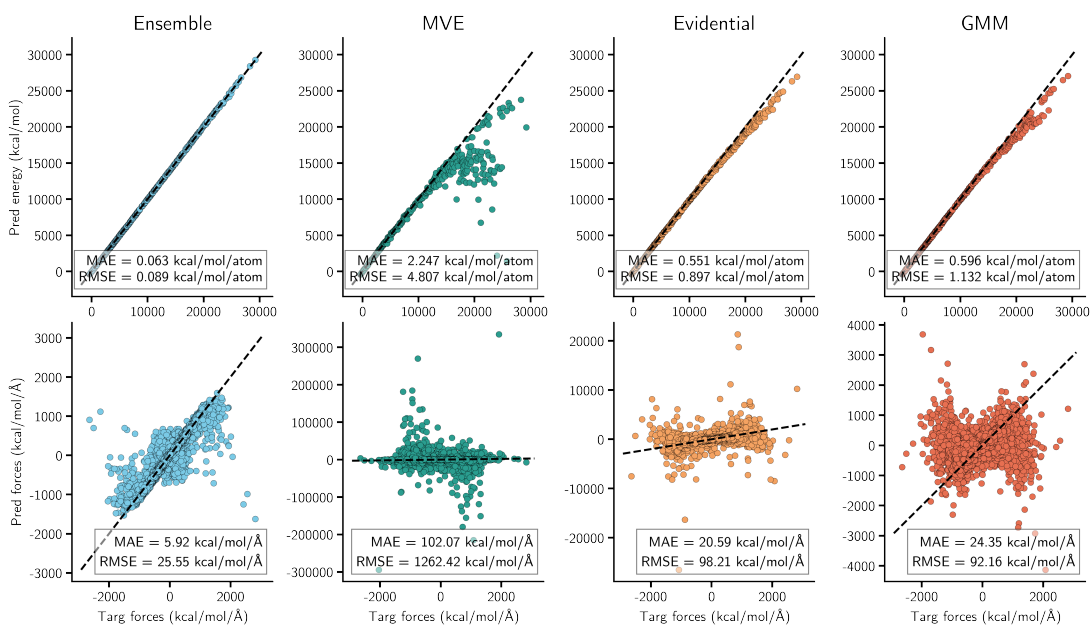

Figure S19. Parity plot showing predicted energy and forces on the test set of silica using the models trained in generation 2 (trained on 1590 initial training data and 100 adversarial samples).

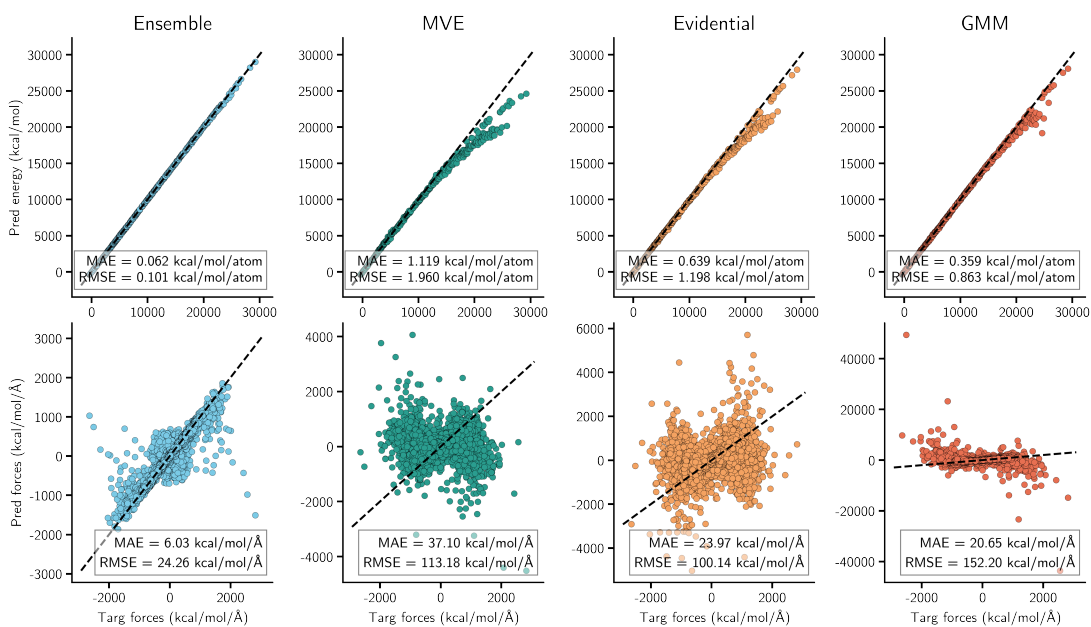

Figure S20. Parity plot showing predicted energy and forces on the test set of silica using the models trained in generation 3 (trained on 1590 initial training data and 200 adversarial samples).

22



\end{document}